\pdfoutput=1
\documentclass[11pt]{article}
\usepackage[dvipsnames]{xcolor}
\usepackage[]{acl}

\usepackage{times}
\usepackage{latexsym}
\usepackage[T1]{fontenc}
\usepackage[utf8]{inputenc}
\usepackage{microtype}
\usepackage{inconsolata}
\usepackage{booktabs}
\usepackage{tabularx}
\usepackage{graphicx}
\usepackage{cleveref}
\usepackage{paralist}

\usepackage{tikz}
\usetikzlibrary{backgrounds, positioning, fit, calc}
\usepackage{pgfplots}
\usepackage{tikzpeople}
\usepackage{tikzsymbols}
\usepackage{twemojis}

\newcommand{\SDprompts}{\texttt{SD}}
\newcommand{\Pprompts}{\texttt{P}}
\newcommand{\Nprompts}{\texttt{N}}

\title{Which Demographics do LLMs Default to During Annotation?}

\author{%
  Johannes Schäfer,
  Aidan Combs,
  Christopher Bagdon,
  Jiahui Li,\\\bfseries
  Nadine Probol,
  Lynn Greschner,
  Sean Papay,
  Yarik Menchaca Resendiz,\\\bfseries
  Aswathy Velutharambath,
  Amelie Wührl,
  Sabine Weber, \and
  Roman Klinger\\
  Fundamentals of Natural Language Processing, University of Bamberg, Germany\\
  \texttt{roman.klinger@uni-bamberg.de}}

\begin{document}
\maketitle

\begin{abstract}
  Demographics and cultural background of annotators influence the
  labels they assign in text annotation -- for instance, an elderly
  woman might find it offensive to read a message addressed to a
  ``bro'', but a male teenager might find it appropriate. It is
  therefore important to acknowledge label variations to not
  underrepresent members of a society. Two research directions
  developed out of this observation in the context of using large
  language models (LLM) for data annotations, namely (1) studying
  biases and inherent knowledge of LLMs and (2) injecting diversity in
  the output by manipulating the prompt with demographic information.
  We combine these two strands of research and ask the question to
  which demographics an LLM resorts when no demographics is
  given. To answer this question, we evaluate which attributes of
  human annotators LLMs inherently mimic.  Furthermore, we compare
  non-demographic conditioned prompts and placebo-conditioned prompts
  (e.g., ``you are an annotator who lives in house number 5'') to
  demographics-conditioned prompts (``You are a 45 year old man and an
  expert on politeness annotation. How do you rate \{instance\}'').
  We study these questions for politeness and offensiveness
  annotations on the \textsc{Popquorn} data set, a corpus created in a
  controlled manner to investigate human label variations based on
  demographics which has not been used for LLM-based analyses so far.
  We observe notable influences related to gender,
  race, and age in demographic prompting, which contrasts with
  previous studies that found no such effects.
\end{abstract}

\section{Introduction}
\colorlet{outerblockscolor}{Cerulean!25}%
\colorlet{innerblockscolor}{Cerulean!10}%
\colorlet{bmcolor}{Orange!25}%
\colorlet{mmcolor}{Orchid!12.5}%
\colorlet{mm2color}{GreenYellow!25}%
\begin{figure}
	\centering
	\begin{tikzpicture}[]
		\node (block1) [rectangle, minimum width=.1\linewidth, text width=.0925\linewidth, inner sep=0pt, minimum height=2.75cm, align=center, anchor=north, text depth = 2.5cm] {\small\textsf{Input}};
		\node (innerblock0) [rectangle, fill=innerblockscolor, minimum width=.15\textwidth, text width=.145\textwidth, below of=block1, node distance=1cm, inner sep=2pt, align=justify, font=\small] {\textsf{Example:\\~\\\textit{If it was an American building it would collapse into itself before the fire spread to more than a few floors.}}};
		
		\begin{pgfonlayer}{background}
			\node (box1) [fit=(block1)(innerblock0), draw=none, fill=outerblockscolor, inner sep=2pt,] {};
		\end{pgfonlayer}
		
		\draw[->] (box1.east)++(0, 1cm) --      ++(2.6cm, 0);
		\draw[->] (box1.east)++(0, -.25cm) --   ++(2.6cm, 0);
		\draw[->] (box1.east)++(0, -1.7cm) -- ++(2.6cm, 0);
		
		\node [rectangle, minimum width=.1\linewidth, text width=.27\linewidth, right=of box1, font=\small, yshift=1.85cm, xshift=-.915cm] {\centering\textsf{Annotation}\\\textsf{(Offensiveness)}};
		\node (ann) [rectangle, minimum width=.5cm, right=of box1, xshift=0cm, yshift=2cm] {};
		\node (p1) [builder,minimum size=.75cm, right=of box1,yshift=.9cm,xshift=-1mm]{};
		\node (p2) [graduate,female,minimum size=.75cm, right=of box1,yshift=-.35cm,xshift=-1mm]{};
		\node (llm) [rectangle, fill=mmcolor, minimum width=1.25cm, text width=.15\linewidth,align=center, right=of box1, xshift=-.35cm, yshift=-1.6cm, inner sep=1pt, font=\small] { \twemoji[height=7.5mm]{robot} \\ \textsf{LLM}};
		
		\begin{pgfonlayer}{background}
			\node (box2) [fit=(ann)(p1)(p2)(llm), inner sep=3pt, draw, dashed, minimum width=2.155cm,] {};
		\end{pgfonlayer}
		
		\node [rectangle, xshift=3.cm, yshift=1cm] at (box1.east) { \dSadey[2][red] };
		\node [rectangle, xshift=3.cm, yshift=-.25cm] at (box1.east) {\dSmiley[2][green]};
		\node (qm) [rectangle, xshift=2.75cm, yshift=-1.7cm, font=\huge] at (box1.east) {?};
		
		\draw[->] (qm.east)++(0, .1cm) --      ++(.3cm, .3cm);
		\draw[->] (qm.east)++(0, -.1cm) --      ++(.3cm, -.3cm);
		
		\node (a1) [rectangle, xshift=.7cm, yshift=.4cm] at (qm.east) { \dSadey[2][red]};
		\node (a2) [rectangle, xshift=.7cm, yshift=-.4cm] at (qm.east) {\dSmiley[2][green]};
		
		\node (p12) [builder,minimum size=.45cm, right=of a1, xshift=-.75cm]{};
		\node at (p12.center) [xshift=-.37cm, font=\huge] {(}; 
		\node at (p12.center) [xshift=.37cm, font=\huge] {)}; 

		\node (p22) [graduate,female,minimum size=.45cm, right=of a2, xshift=-.75cm]{};
		\node at (p22.center) [xshift=-.37cm, font=\huge] {(}; 
		\node at (p22.center) [xshift=.37cm, font=\huge] {)}; 
		
	\end{tikzpicture}
	\caption{Our objective is to identify which human demographic groups are mimicked by LLMs during subjective annotation tasks on text data.}
	\label{fig:overview}
\end{figure}%
In some text\footnote{We provide our code and model predictions on \url{https://www.uni-bamberg.de/en/nlproc/resources/llms-default-demographics/}.} annotation tasks, it is feasible to obtain an aggregated
ground truth label, for instance in named entity annotation
\citep{yadav-bethard-2018-survey} or semantic role labeling
\citep{shi-etal-2020-semantic}. In other tasks, perhaps even in the
majority of tasks, it is more obvious that annotators' traits
influence the label assignments, for instance in sentiment annotation
\citep{Liu2012}, emotion annotation
\citep{klinger-2023-event,plaza-del-arco-etal-2024-emotion}, or
personality profiling \citep{neuman-2015-personality}. The diversity
of annotations, conditioned on annotators' profiles, has been
recognized as an important variable to consider, instead of
aggregating all labels to an adjudicated score, which might not
correspond to any of the annotators \citep{plepi-etal-2022-unifying}.

With the advent of (instruction-tuned) large language models (LLMs),
automatic data annotation and zero or few shot predictions became more
popular \citep{brown2020languagemodelsfewshotlearners}. However,
language models do not provide the same diversity as human annotators
do in a simple zero-shot setup, which can lead to a
lower performance of the labeling process
\citep{bagdon-etal-2024-expert}. To mitigate this problem, it is
important to understand whether large language models exhibit a
default persona when acting as annotators and how this can be
controlled to ensure greater diversity in annotations.
One concern is that large language models may reflect biases present in their training data, which can disproportionately emphasize certain viewpoints.
This raises the issue of potential marginalization of minority perspectives in the outputs generated by these models, making it crucial to address this bias for equitable representation in annotations.
One idea to
address this limitation is socio-demographic prompting
\citep{muscato-etal-2024-overview}, where we guide the model by specifying characteristics in the prompt,
like age, gender, ethnicity or socio-economic status.

Previous research has explored the effects of using prompts informed
by demographic or cultural contexts on model predictions, but did not
find consistent patterns \citep{beck-etal-2024-sensitivity,
  mukherjee2024culturalconditioningplaceboeffectiveness}.  We build on
top of this work and particularly contribute in two directions of
socio-demographic prompting. We study if
large language models default to a particular demographic, i.e., we
evaluate if the LLM-based predictions more closely align with those of people of a particular
demographic when not being conditioned. 
Our general objective is depicted in \Cref{fig:overview}.
Additionally, we utilize placebos as suggested by \citet{mukherjee2024culturalconditioningplaceboeffectiveness} -- irrelevant information that should not affect the model's output -- to compare the impact of demographic prompts and evaluate the stability of the model's predictions.
More concretely, we answer the following 
questions.

\newcommand{\RQoneText}{What default demographic values can we infer from the annotation behavior of large language models?}
\newcommand{\RQtwoText}{Are the changes to the models' annotations more pronounced for demographic prompting than with non-relevant additional information to the prompt?}
\newcommand{\RQthreeText}{How do task properties of offensiveness rating vs.\ politeness rating influence the role of demographic information in prompts?}
\newcommand{\RQfourText}{Are observed patterns consistent across different large language models?}
\begin{compactitem}
\item [RQ1] \RQoneText
\item [RQ2] \RQtwoText
\item [RQ3] \RQthreeText
\item [RQ4] \RQfourText
\end{compactitem}

The remainder of this paper is structured as follows: We review related
work in \Cref{sec:relatedwork} and explain our data set and task choices as well as
prompt setup in \Cref{sec:exp}. In \Cref{sec:results}, we present the
results of our experiments and conclude including a discussion of possible
future work in \Cref{sec:conclusion}.

\section{Related Work}
\label{sec:relatedwork}
This paper relates to various areas which we review in the following,
namely biases in large language models, perspectivism, and
(socio-demographic) prompting of LLMs.

\subsection{Inherent Knowledge and Biases in LLMs}
Large language models may be understood as databases that
store information encoded in the training data
\citep{petroni-etal-2019-language}. These data are stored in a
probabilistic manner, which allows models to show sometimes
unexpected generalization capabilities beyond the original
instructions available in the training data, i.e., emergent abilities
\citep{wei2022emergent}.
It has also been shown that prompting models can adapt parameters during inference, similar to how backpropagation works \citep{vonoswald2023transformerslearnincontextgradient}.
Therefore, it is reasonable to view prompting as a form of programming, as it involves manipulating model behavior by crafting specific inputs to achieve desired outputs \citep{Beurer2024}.

There are cases in which the information requested from a large
language model is not available in the training data ``as is'' and the
generalization abilities reach its limits. In such cases, the model
might output text that is not correct. Such cases are sometimes
considered ``hallucinations''. The confidence of hallucinations
is often lower than for correct information
\citep{farquhar_detecting_2024}.

In cases in which subjective properties are requested, the models
therefore rely on ``emergent'' abilities. There is some research in
understanding the ``stances'' or ``worldviews'' encoded in language
models. For instance, \citet{Motoki2023,feng-etal-2023-pretraining} study
political biases and how these lead to unfair
models. \citet{ceron2024promptbrittlenessevaluatingreliability} study
the reliability and robustness of such
stances. \citet{wright2024revealingfinegrainedvaluesopinions} extend
such analysis to more fine-grained opinions. The opinions expressed by
models correspond more to particular subsets of a society than to
others \citep{Santurkar2023}.

\subsection{Perspectivism}
Natural language processing for a long time focused on seemingly
objective tasks such as parsing or named entity
recognition. Therefore, it has been a standard approach to adjudicate
annotations into a single gold standard \citep{stubbs-2011-mae}. To do
so, a set of methods for aggregation that considers the distributions
and disagreements have been developed
\citep{paun-etal-2018-comparing}. More recently, research moved
towards acknowledging the importance of
disagreements. \citet{plank-2022-problem} has been one of the first to
discuss challenges and potential approaches to this problem. Recent
work looked into methods to consider disagreements
\citep{fleisig-etal-2024-perspectivist} and discusses ethical
considerations \citep{valette-2024-perspectivism}. Nowadays, this
field perspectivism developed
dedicated shared tasks \citep{uma-etal-2021-semeval} and workshops
\citep{nlperspectives-2024-perspectivist}.

Disagreement is related to confidence of annotators (and models), an
aspect that has received some
attention. \citet{baumler-etal-2023-examples} made use of this
property in the context of active learning, to understand which
instances require annotations by multiple people in order to understand the
disagreements. \citet{troiano-etal-2021-emotion} and \citet{baan-etal-2024-interpreting}
showed that annotators' own confidence predicts
inter-annotator agreement scores.

To understand disagreement in labeling better, more and more corpora are being published with more detailed information about the annotators.
\citet{troiano-etal-2023-dimensional} annotated event
reports for emotions and appraisals and collected demographic
information, personality, and current emotional state of both the person
who lived through the event and multiple annotators that read the
event description. \citet{plepi-etal-2022-unifying} studied the role
of demographics, automatically extracted from the data, on the
perception of social norms. \citet{bizzoni-etal-2022-predicting}
discuss the role of individual differences on the judgement of
literary quality. \citet{romberg-2022-perspective} integrates
perspectivism in argument mining, by making explicit the subjective
nature of argument interpretation. \citet{may-etal-2024-perspectivist}
study the effect of demographics on the role of numbers in social judgements.  \citet{frenda-etal-2023-epic} study how the perception of
irony varies by nationality, employment status, student status,
ethnicity, age, and gender. \citet{sachdeva-etal-2022-measuring}
measure different aspects of hate speech which include sentiment, disrespect,
insult, attacking/defending, humiliation, inferior/superior status,
dehumanization, violence, genocide, and a 3-valued hate speech
benchmark label. They study these variables under the condition of
identity target groups and annotator
demographics. \citet{xu-etal-2023-dissonance} look into disagreement
of legal case outcome differences. Next to subjective tasks,
perspectivism has also been considered in seemingly objective tasks,
for instance in named entity recognition
\citep{peng-etal-2024-different} and natural language inference
\citep{gruber-etal-2024-labels}.

The large \textsc{Popquorn} corpus \citep{pei-jurgens-2023-annotator}
has been created specifically to study perspectivism of annotators and
the relationship between demographics and annotations in the tasks of offensiveness detection, question answering,
text rewriting and style transfer, and politeness rating.  We use this corpus because it has been created
for the study of perspectivism, but it has not yet been used to analyze large language models.

\subsection{Prompting for Automatic Data Set Annotation or Zero-Shot
  Predictions}
While fine-tuning models is currently still the state-of-the-art
approach to obtain the best possible performance for a variety of
natural language processing tasks, prompting language models for a
zero-shot of few-shot prediction gained popularity recently. This is
due to the possibility of efficiently adapting model outputs by prompt
engineering, without fine-tuning the model -- in fact, prompt
optimization (by a human or automatically) can be seen as
parameter-efficient model adaptation.

This field builds on top of instance-based classification methods and
became popular with the work by \citet{NIPS2009_1543843a} who suggest
to perform dataless classification by semantically encoding output
concepts. Another non-NLP example is \citet{Banerjee2022} who embed
emotion concepts for zero-shot classification of body
gestures. A milestone in the natural language processing community is
the work by \citet{yin-etal-2019-benchmarking} who show how natural
language inference can be applied across multiple classification tasks
and \citet{brown2020languagemodelsfewshotlearners} who show that
auto-regressive language models are zero-shot learners.

Since then, a set of studies have been proposed, including work that
focuses on cross-linguality \citep{Bareiss2024}, data augmentation
\citep{chen-shu-2023-promptda}, emotions
\citep{plaza-del-arco-etal-2022-natural,bagdon-etal-2024-expert},
named entity recognition \citep{shen-etal-2023-promptner}, and
sentiment classification
\cite{fei-etal-2023-reasoning,ma-etal-2022-prompt}. A more
comprehensive survey has been provided by
\citet{li-2023-practical}. Prompts can also be learned, but this setup
is out of scope for our work in this paper
\citep{liu2021pretrainpromptpredictsystematic}.

\subsection{Socio-Demographic Prompting}
Automatic annotation with language models is not a replacement for
human annotation. Humans have previous world knowledge, experiences,
and perspectives on a matter that differ individually. LLMs have difficulties replicating these differences, causing issues in annotation. For instance,
\citet{bagdon-etal-2024-expert} show that the diversity of
annotations that is beneficial in comparative annotations by multiple
people cannot be straightforwardly replaced by multiple runs of a
language model. \citet{lee-etal-2023-large} focus on this aspect in particular and
find poor alignment of the distribution of labels predicted by LLMs with 
human annotations on natural language inference tasks.

Most relevant for our work are the following
studies. \citet{beck-etal-2024-sensitivity} study the impact of
demographic information on subjective annotation tasks. They find that
model variation is larger across other parameters such as prompt formulation techniques than demographic information (e.g., age and gender) in the
prompt. 
\citet{mukherjee2024culturalconditioningplaceboeffectiveness} examine cultural aspects like food preferences and find that most language models exhibit significant response variability, casting doubt on the reliability of socio-demographic prompting.
To understand if the variables influence the
annotation in a systematic way, they compare the predictions to what
they call ``placebos'' -- information that should not influence the
prediction but looks like relevant parameters (favorite planet or
house number).

Additionally, \citet{sun-etal-2025-sociodemographic} highlight that most LLMs show demographic biases in subjective judgment tasks, favoring perceptions from White participants over those from Asian or Black participants.
\citet{hu-collier-2024-quantifying} find that incorporating persona variables in LLM prompting improves model predictions slightly, especially in conditions of significant annotator disagreement.
\citet{movva-etal-2024-annotation} reveal that while GPT-4 shows reasonable alignment with human assessments of safety, there are significant demographic disparities in how well it correlates with different annotator groups.

Finally, there is a set of studies that investigate biases on large language models
\citep[i.a.]{cheng-etal-2023-marked,santy-etal-2023-nlpositionality}.
Their findings suggest that these models may generate outputs reflecting racial stereotypes and exhibit some performance disparities across different demographic groups, indicating potential biases in their design and outputs.

Our work combines aspects of previous research, namely on \textit{biases}, \textit{placebos},
and \textit{demographics}.

\section{Experimental Setting}
\label{sec:exp}
This section presents the methodology and resources utilized in our experiments.

\subsection{Data Sets}
We chose the \textsc{Popquorn} data for our experiments
\citep{pei-jurgens-2023-annotator}. In contrast to data used by other
previous research we are aware of, these data have been particularly
sampled for the study of annotators' properties and their impact on
annotation tasks. Therefore, these data render themselves as a
straightforward choice also for an LLM-based analysis. 
The original data stem from the Ruddit corpus
\citep{hada-etal-2021-ruddit} which was originally annotated via
best--worst scaling. This may lead to different label frequencies than
rating scale annotations and is considered as not appropriate for skewed
distributions \citep{Louv-Food-BWS,Louv-BWS}. The authors of \textsc{Popquorn}
reannotated the data with 1--5 rating scales, which we adopt for our
annotation setup. Further, it is noteworthy that the creators of the
data set removed annotators with particularly low annotation
competency according to \textsc{Mace} \citep{hovy-etal-2013-learning},
to filter out potential random noise.

The \textsc{Popquorn} data set consists of 45,000 annotations from
1,484 annotators with information on their gender, race,
age, occupation and education. We use data from the subjective partitions for the offensiveness and politeness rating tasks.
To equalize the number of annotations per instance, we randomly sample three
annotations for each instance.%
\footnote{%
	The decision to retain three annotators per instance was a deliberate choice to standardize our methodology, as the original dataset has varying numbers of annotators.
	We aimed to ensure consistency across our analysis and view the balance we achieve as a necessity for the comparison to non-demographic prompting, which always produces one prediction per instance.
	With our method, we avoid a skewed distribution towards instances with a larger number of annotators in demographic prompting.
	We have checked whether the distribution is skewed due to our sampling:
	For the offensiveness data, our sample consists of 35\% of the overall annotations from \textsc{Popquorn}.
	As for the distribution of each demographic attribute in our sample: It comprises 31\%-38\% of the annotations per attribute (standard deviation: 1\%).
	For the politeness data, our sample consists of 45\% of the overall annotations.
	As for the distribution of each demographic attribute in our sample: It comprises 34\%-56\% of the annotations per attribute (standard deviation: 4\%; the wide range is due to the fact that some attributes being extremely rare in the original dataset, leading to outliers).
	Our sample maintains a certain degree of representativeness of the demographics in \textsc{Popquorn}.} %
When sampling, we omit the annotations
where the annotator preferred not to disclose demographic information.
We exclude data instances with fewer than three annotations
remaining. The two subsets of the sample which we use in our
experiments consist of 4,500 annotations of 1,500 instances for
offensiveness ratings and 11,151 annotations of 3,717 instances for
politeness ratings.

\begin{table*}[t]
	\centering\small
	\begin{tabularx}{\linewidth}{p{1.8cm}p{8.1cm}X}
		\toprule
		Prompt Type & Prompt 1 Template & Prompt 2 Template \\
		\cmidrule(r){1-1}\cmidrule(lr){2-2}\cmidrule(l){3-3}
		
		Socio-Demographic (\SDprompts) & You are a person of gender [gender], race [race],
		age [age], occupation [occupation],  education [education]. Given a text,
		how do you rate the [task] of the following text? Possible
		values are [task-values]. In your answer, only provide the label you chose. 
		& Imagine [Prompt 1 with initial character in lower case] \\  
		
		\cmidrule(r){1-1}\cmidrule(lr){2-2}\cmidrule(l){3-3}
		
		Placebo (\Pprompts) & You are a person of height [height], Zodiac sign [zodiac sign], living in a house with the number [house number], enjoying [enjoying], and whose favorite color is [favorite colour].  Given a text, how do you rate the [task] of the following text? Possible values are [task-values].
		In your answer, only provide the label you choose.
		& Imagine [Prompt 1 with initial character in lower case] \\  
		
		\cmidrule(r){1-1}\cmidrule(lr){2-2}\cmidrule(l){3-3}
		Non-Demographic (\Nprompts)& You are a person of any gender, any race, any age, with any occupation, and any education.
		Given a text, how do you rate the [task] of the following text?
		Possible values are [task-values]. 
		In your answer, only provide the label you choose. & Given a text, how do you rate the [task] of the following text?
		Possible values are [task-values]. 
		In your answer, only provide the label you chose. \\
		\bottomrule
	\end{tabularx}
	\caption{Prompt types and Templates for system messages. For the [task] offensiveness, the [task-values] are `not offensive', `slightly offensive', `moderately offensive', `very offensive' or `extremely offensive'. For the [task] politeness, the [task-values] are `not polite', `slightly polite', `moderately polite', `very polite' or `extremely polite'.}
	\label{tab:prompts}
\end{table*}

\begin{table}[t]
	\centering\small
	\begin{tabularx}{\linewidth}{lX}
		\toprule
		Placebo Attribute & Possible Values \\
		\cmidrule(r){1-1}\cmidrule(l){2-2}
		Height & 140 cm, 150 cm, 160 cm, 170 cm, 180 cm, 190 cm, 200 cm, 210 cm.\\
		Zodiac sign & Aries, Taurus, Gemini, Cancer, Leo, Virgo, Libra, Scorpio, Sagittarius, Capricorn, Aquarius, Pisces.\\
		House number & 6, 12, 13, 24, 45, 68, 98, 122, 234, 1265.\\
		Enjoying & food, sleep, friends. \\
		Favorite colour & red, green, blue, yellow, purple, turquoise, orange, pink, black, white, brown. \\
		\bottomrule
	\end{tabularx}
	\caption{Sets of values for placebo attributes used in \Pprompts~prompts in our experiments.}
	\label{tab:placebovalues}
\end{table}

\subsection{Prompt Setup}\label{sec:promptsetup}
We use three different prompt setups, namely socio-demographically
informed (\SDprompts) prompts, placebo-informed (\Pprompts) prompts,
and prompts without additional information (\Nprompts\ prompts). We
show the templates used for these prompts in \Cref{tab:prompts}. The input
to the LLM consist of a concatenation of the system message with the 
respective text to be classified. The values for the demographic attributes 
are taken from the demographic data of the annotators included in 
the sample.  The sets of values for the placebo attributes are displayed in
\Cref{tab:placebovalues}. During prompting, each placebo attribute
value is randomly sampled from the respective set.

\subsection{Model Choice and Access}
We use two LLMs, namely
GPT-4o \citep{gpt4o} and
Claude \citep{claude}.  We access
both models via their respective APIs with default hyperparameters.  The total cost for using
GPT-4o was \$54 and the cost for using Claude was \$60.

\subsection{LLM Output Parsing}
We transform the texts generated by the LLMs for the annotation tasks into a defined label set
of five categories, corresponding to the 1--5 rating scale for each
of the two tasks.  We use the Langchain
StrOutputParser\footnote{\url{https://api.python.langchain.com/en/latest/output_parsers/langchain_core.output_parsers.string.StrOutputParser.html}}
to interpret the output from GPT-4o as one of the designated labels,
while we use the output from Claude directly.

This process does not successfully parse the output in all cases.
We encountered 22 error cases with GPT-4o and three with Claude.
Of the failed cases with GPT-4o, 14 were due to the input text from the original data set being in Polish instead of English.
This resulted in the model predicting labels in Polish, despite the prompt specifying an English-only label set.
The remaining eight cases involved instances with very short text or questions, which the model misinterpreted.
In these cases, GPT-4o asked for further input instead of performing the intended classification task.

For Claude, the three instances of failure involved the names of famous actors in the text.
This triggered the model’s copyright protection protocols, preventing it from reproducing or paraphrasing potentially copyrighted content.

Given the total number of instances analyzed in our experiments, this failure rate is insubstantial.
Consequently, we disregard the output in the cases where parsing was unsuccessful and assign the labels ``not offensive'' or ``not polite''.\footnote{Our decision to label such instances with the negative class, rather than
	removing them, was made to maintain comparability across different experimental settings. Removing instances due
	to a single failed label would distort the consistency of the number of variants of prompts per text instance.}

\subsection{Evaluation Settings}
\label{sec:evalsetting}
We conduct various analyses of the labels generated by the LLMs, utilizing the different prompt setups described in \Cref{sec:promptsetup}. 
First, we examine the extent to which LLM predictions align with the judgments of different human annotators by comparing the outputs generated from non-demographic (\Nprompts) prompts to the human annotations available for each instance.
Second, we examine the impact of socio-demographic prompting by comparing the models' automatic annotations produced with \SDprompts\ prompts to those produced with \Nprompts\ prompts.
In this analysis, we assess the differences in the LLMs' annotations when demographic data is included versus when it is omitted.
Thirdly, we investigate the effect of placebo prompting in a similar manner by comparing the automatic annotations generated with \Pprompts\ prompts to those created with \Nprompts\ prompts.

\section{Results}
\label{sec:results}

In this section, we present the results of our experiments based on
the annotations generated by the LLMs for two labeling tasks. Each of
our research questions is addressed with particular results derived
from the various experiments conducted. Detailed results for
individual prompts show that the two
\Nprompts\ prompts behave similarly (see \Cref{sec:appind}). We also
observed this for the N and SD prompt templates.
Thus, we here report all results as an average over the two respective
prompt templates for the different prompt types.

\begin{table}[t]
  \centering \small \setlength{\tabcolsep}{3pt}
  \begin{tabularx}{\linewidth}{X rr rr}
    \toprule
    &\multicolumn{2}{c}{Offensiveness}&\multicolumn{2}{c}{Politeness} \\ 
    \cmidrule(r){2-3}\cmidrule(l){4-5}
    Socio-Demographic Attribute & GPT-4o & Claude & GPT-4o  & Claude  \\
    \cmidrule(r){1-1} \cmidrule(lr){2-2} \cmidrule(lr){3-3} \cmidrule(lr){4-4} \cmidrule(l){5-5}
	\textbf{Age}                       & **0.01   & **0.01  & 0.00    & 0.00    \\
	\multicolumn{5}{l}{\textbf{Gender (ref.: Male)} } \\
	\hspace{1em}Female                            & 0.00    & $-$0.03  & $-$0.05   & $-$0.05   \\
	\hspace{1em}Non-binary                        & $-$0.06   & $-$0.01  & $-$0.05   & $-$0.05   \\
	\multicolumn{5}{l}{\textbf{Race (ref.: White)}} \\
	\hspace{1em}Asian                             & 0.09    & 0.03   & $-$0.08   & 0.00    \\
	\hspace{1em}Black/Afri.\ Am.         & ***0.22 & **0.19  & **0.14   & **0.15   \\
	\hspace{1em}Hispanic or Latino                & $-$0.11   & $-$0.05  & 0.09    & 0.12   \\
	\hspace{1em}\textit{Other race}              & $-$0.14   & $-$0.26 & $-$0.17   & $-$0.15   \\
	\multicolumn{5}{l}{\textbf{Occupation (ref.: Employed)}} \\
	\hspace{1em}Unemployed                        & 0.04    & $-$0.08  & $-$0.08   & $-$0.08   \\
	\hspace{1em}Homemaker                         & $-$0.07   & $-$0.02  & $-$0.03   & $-$0.06   \\
	\hspace{1em}Retired                           & $-$0.11   & $-$0.13  & 0.03    & 0.03    \\
	\hspace{1em}Self-employed                     & 0.04    & 0.02   & $-$0.03   & $-$0.03   \\
	\hspace{1em}Student                           & 0.13    & 0.13   & $-$0.11   & $-$0.13   \\
	\hspace{1em}\textit{Other occupation}                  & $-$0.05   & 0.02   & 0.08    & 0.11    \\
	\multicolumn{5}{l}{\textbf{Education (ref.: Less than high school)}} \\
	\hspace{1em}High school dipl. & $-$0.01   & $-$0.08  & $-$0.34  & *$-$0.37  \\
	\hspace{1em}College degree                    & 0.05    & $-$0.09  & *$-$0.43  & *$-$0.48  \\
	\hspace{1em}Graduate degree                   & 0.06    & $-$0.01  & *$-$0.36  & *$-$0.44  \\
	\hspace{1em}\textit{Other education}                   & $-$0.02   & 0.01   & *$-$0.50  & **$-$0.57 \\
    \bottomrule
  \end{tabularx}
  \caption{Coefficients indicating the effect of particular human
    demographic categories on the distance between human and LLM
    annotations, calculated using mixed-effects regression models with random intercepts for annotators and instances. Statistical significance is calculated using standard error (see \Cref{sec:appful} for these values) and is here marked by asterisks: * corresponds to P $\leq$ 0.05, ** to P $\leq$ 0.01, and *** to P $\leq$ 0.001.}
  \label{tab:def}
\end{table}

\subsection{RQ1: \RQoneText}
We approach the identification of the default persona of the models in two
different ways following the first two settings as described in
\Cref{sec:evalsetting}. The following sections present and discuss
the results of these analyses.

\noindent\textbf{Which socio-demographic
attributes of human annotators does an LLM inherently mimic in the absence
of explicit information (\Nprompts\ prompts)?}
\Cref{tab:def} shows the results for this analysis which corresponds
to the first setting described in \Cref{sec:evalsetting}. We combine socio-demographic
attributes that are represented in only a few cases. The table
specifies the reference categories (ref.) used for each of the
categorical variables.\footnote{
	The reference categories for the categorical variables occupation and education are chosen according to the recommendations of \citet{doi:10.1177/0081175020982632}.
	For race and gender, their scheme does not result in unambiguous recommendations, so we choose the categories we expect to lie closest to the models' defaults: male for gender and white for race.}

\Cref{tab:def} shows the coefficients of regression models predicting absolute values of the distance between LLM annotations (\Nprompts\ prompts) and annotations provided by human annotators for the same instance. Independent variables are the socio-demographic characteristics of the human annotators. We report results separately for the two models (GPT-4o and
Claude) and two classification tasks (offensiveness and politeness rating). The reported coefficients can be
interpreted as the effect of particular human demographic categories
on the distance between human and LLM annotations (change in distance between the category in question and the reference category).
Positive coefficients indicate that the respective LLM is further from the human
annotators in that demographic category than human annotators in the reference category, i.e., less accurate.  From this we can follow that the
categories with positive, statistically significant coefficients are
those categories which the respective model does not default to.
In contrast, negative coefficients indicate that
the LLM is closer to the human annotators in the given demographic category than it is to the human annotators in
the reference category; i.e., more accurate. Categories with negative and statistically significant coefficients are those
that are nearer to the model’s default than the reference category.

The analysis of coefficients (see \Cref{tab:def}) reveals significant biases in the
LLMs' predictions based on demographic factors, particularly
concerning race and age. Specifically, the models demonstrate a
tendency to align more closely with annotations provided by persons
identifying as White as opposed to those identifying as Black or African
American. The distance between their annotations measures .19 to
.22 Likert scale points for offensiveness rating and .14 to .15
points for politeness rating. Additionally, the LLMs are
progressively less accurate in reflecting the views of older
individuals at offensiveness rating, with a .01 point increase in
distance for each year of age. In terms of educational background, the
models exhibit a greater discrepancy from those with less than a high
school education compared to those with higher educational levels,
ranging from .34 to .57 points for politeness rating. One possible explanation for these
discrepancies by annotator sociodemographics is that the models' training data may lack
representation for these demographic groups.

Notably, this analysis does not show any significant effects related to gender
or occupational status. This could be because the LLMs do not consistently favor any particular group of people on these dimensions, or it could be because there are in fact no systematic differences in human annotation along these dimensions. Overall, while the statistically significant
differences in annotation distances suggest some demographic biases,
we emphasize that these effects are small compared to the full Likert scale range. In addition, the R2 Marginal values for these models (reported in Appendix~\ref{sec:appful}) are quite low, indicating that the sociodemographic categories explain little of the variability in human--LLM annotation differences.
This complexity makes it challenging to establish a clear default
persona for the LLMs. Therefore, we perform a second analysis in the
following.

\definecolor{ourgray}{gray}{0.6}
\begin{table*}[t]
	\centering \setlength{\tabcolsep}{5.75pt}
	\small
	\renewcommand{\arraystretch}{.975}
	\begin{tabular}{ll rrr rrr}
		\toprule
		&& \multicolumn{3}{c}{Offensiveness} & \multicolumn{3}{c}{Politeness} \\
		\cmidrule(r){3-5} \cmidrule(r){6-8}
		\multicolumn{2}{l}{Socio-Demographic Attribute}& Count&$\Delta_\mu$ (GPT-4o)&$\Delta_\mu$ (Claude)& Count&$\Delta_\mu$ (GPT-4o)&$\Delta_\mu$ (Claude) \\
		\cmidrule(r){1-2} \cmidrule(rl){3-3} \cmidrule(lr){4-4} \cmidrule(r){5-5} \cmidrule(rl){6-6} \cmidrule(rl){7-7} \cmidrule(l){8-8}
		\multicolumn{2}{l}{Gender} & & & & & & \\
		& Male        & 2,157 & 0.18 & 0.17 & 5,195 & 0.26 & 0.20 \\
		& Female      & 2,219 & 0.20 & 0.15 & 5,623 & 0.24 & 0.22 \\
		& Non-binary  & 124  & 0.29 & 0.17 & 333  & 0.24 & 0.17 \\
		\multicolumn{2}{l}{Race} & & & & & & \\
		& White                     & 3,396 & 0.18 & 0.16 & 8,163 & 0.25 & 0.20 \\
		& Hispanic or Latino        & \textcolor{ourgray}{95}   & \textcolor{ourgray}{0.21} & \textcolor{ourgray}{0.11} & 790  & 0.22 & 0.21 \\
		& Native American           & \textcolor{ourgray}{97}   & \textcolor{ourgray}{0.25} & \textcolor{ourgray}{0.12} & \textcolor{ourgray}{0}    & \textcolor{ourgray}{--}   & \textcolor{ourgray}{--}    \\
		& Arab American             & \textcolor{ourgray}{17}   & \textcolor{ourgray}{0.32} & \textcolor{ourgray}{0.06} & \textcolor{ourgray}{0}    & \textcolor{ourgray}{--}   & \textcolor{ourgray}{--}    \\
		& Native Hawaiian or Pacific Islander & \textcolor{ourgray}{0}  & \textcolor{ourgray}{--}   & \textcolor{ourgray}{--}   & \textcolor{ourgray}{36}   & \textcolor{ourgray}{0.31} & \textcolor{ourgray}{0.15} \\
		& American Indian or Alaska Native & \textcolor{ourgray}{0}  & \textcolor{ourgray}{--}   & \textcolor{ourgray}{--}   & \textcolor{ourgray}{28}   & \textcolor{ourgray}{0.18} & \textcolor{ourgray}{0.23} \\
		& Black or African American  & 559  & 0.22 & 0.15 & 1,386 & 0.25 & 0.21 \\
		& Asian                     & 336  & 0.21 & 0.15 & 731  & 0.24 & 0.24 \\
		& Hebrew                    & \textcolor{ourgray}{0}    & \textcolor{ourgray}{--}   & \textcolor{ourgray}{--}   & \textcolor{ourgray}{17}   & \textcolor{ourgray}{0.18} & \textcolor{ourgray}{0.26} \\
		\multicolumn{2}{l}{Age} & & & & & & \\
		& 18-24                    & 499  & 0.21 & 0.16 & 1,241 & 0.25 & 0.16 \\
		& 25-29                    & 444  & 0.20 & 0.16 & 894  & 0.25 & 0.18 \\
		& 30-34                    & 540  & 0.16 & 0.15 & 1,190 & 0.24 & 0.17 \\
		& 35-39                    & 532  & 0.20 & 0.19 & 834  & 0.26 & 0.18 \\
		& 40-44                    & 418  & 0.19 & 0.18 & 1,176 & 0.23 & 0.21 \\
		& 45-49                    & 388  & 0.19 & 0.15 & 957  & 0.22 & 0.22 \\
		& 50-54                    & 314  & 0.17 & 0.17 & 995  & 0.27 & 0.21 \\
		& 54-59                    & 627  & 0.18 & 0.15 & 1,083 & 0.26 & 0.22 \\
		& 60-64                    & 251  & 0.22 & 0.15 & 1,193 & 0.24 & 0.22 \\
		& >65                      & 487  & 0.20 & 0.15 & 1,588 & 0.25 & 0.26 \\
		\multicolumn{2}{l}{Occupation} & & & & & & \\
		& Unemployed               & 571  & 0.19 & 0.14 & 1,328 & 0.28 & 0.19 \\
		& Employed                 & 2,189 & 0.19 & 0.17 & 4,944 & 0.24 & 0.20 \\
		& Homemaker                & 199  & 0.24 & 0.14 & 784  & 0.23 & 0.21 \\
		& Retired                  & 500  & 0.19 & 0.15 & 1,783 & 0.26 & 0.25 \\
		& Other                   & \textcolor{ourgray}{86}   & \textcolor{ourgray}{0.16} & \textcolor{ourgray}{0.22} & 268  & 0.28 & 0.19 \\
		& Self-employed            & 617  & 0.17 & 0.17 & 1,395 & 0.23 & 0.21 \\
		& Student                 & 338  & 0.23 & 0.14 & 649  & 0.23 & 0.16 \\
		\multicolumn{2}{l}{Education} & & & & & & \\
		& Less than a high school diploma & \textcolor{ourgray}{84} & \textcolor{ourgray}{0.18} & \textcolor{ourgray}{0.17} & \textcolor{ourgray}{76}   & \textcolor{ourgray}{0.30} & \textcolor{ourgray}{0.22} \\
		& High school diploma or equivalent & 1,379 & 0.19 & 0.14 & 3,312 & 0.26 & 0.17 \\
		& Graduate degree          & 846  & 0.21 & 0.16 & 2,160 & 0.23 & 0.25 \\
		& College degree            & 2,098 & 0.18 & 0.17 & 5,352 & 0.24 & 0.21 \\
		& Other                    & \textcolor{ourgray}{93}   & \textcolor{ourgray}{0.25} & \textcolor{ourgray}{0.11} & 251  & 0.24 & 0.23\\
		\bottomrule
	\end{tabular}
	\caption{Sample sizes (count) and mean distance scores of
          demographic-prompting (\SDprompts\ prompts) predictions in comparison to
          predictions with \Nprompts\ prompts for models GPT-4o and Claude
          at two rating tasks. Variables with very few cases (count~$\leq$~100) are shown
        in gray to indicate a lack or reliability of these numbers.}
	\label{tab:mdist}
\end{table*}

\noindent\textbf{The Effect of Demographic Prompting.}
In a second approach, we examine how the inclusion of demographic
information in the prompt (cf.\ second setting described in
\Cref{sec:evalsetting}) influences the automatic annotation outputs of
the models.  We compare the outputs generated with \SDprompts\ prompts to those with \Nprompts\
prompts. We assess the differences in the LLMs’ annotations with and without the inclusion of demographic data, which enables us to infer the demographics closest to the models' default.

\Cref{tab:mdist} shows the results as differences in the prediction
scores ($\Delta_\mu$).\footnote{Note that there is an overlap in the
  age ranges (50--54, 54--59), a consequence of the survey question
  design in the original annotation task for the \textsc{Popquorn}
  dataset. We have preserved these original categories, despite this
  issue, in order to maintain the fine-grained distinctions they
  provide.} 
For some demographic attributes, there are relatively few samples
(``count'' columns).  This implies that there may not be sufficient
statistical evidence to support the observed average distance values
for certain categories.  We focus our discussion on cases with
count $\geq$ 100.

GPT-4o shows substantial differences in the scores for the
offensiveness task across different gender attribute values: The
prediction differences for Non-Binary (.29) are substantially larger
 than those for Male (.18) and Female (.20). This suggests that
prompting the LLM to act as a non-binary individual has a more
pronounced effect on its predictions, indicating that the male or
female attributes are more aligned with its default persona. This
observation is not consistent across tasks.

Claude shows a notable pattern for the politeness task concerning the
age socio-demographic attribute. With an increasing age, the
prediction difference increases (from .16 to .26).  Similarly, the
occupation attribute reflects the highest difference for the value
Retired.  This indicates that when Claude is prompted to act as an
older individual, it exhibits an increased sensitivity to politeness.

The discrepancy between the results from the analysis displayed in \Cref{tab:def} to those from the analysis displayed in \Cref{tab:mdist} are not a contradiction.
These analyses are based on distinct interpretations of the default persona of the models.
\Cref{tab:def} indicates a poor representation of some demographics in the models' responses.
Conversely,  \Cref{tab:def} highlights that, when prompted with specific demographics, the output was marginally closer to the default, suggesting that the models' interpretation of these demographic attributes influences its behavior only slightly.
Thus, the model exhibits a measurable lack of alignment with certain demographics while it simultaneously demonstrates minimal variation in its behavior when prompted to act as a person with those demographics. 

\subsection{RQ2: \RQtwoText}
We investigate whether the modifications to the model are more substantial for socio-demographic prompting compared to non-relevant additional information provided in the prompts.
Specifically, we analyze how the models' predictions for the studied tasks are affected when presented with placebo information (cf.\ third setting from \Cref{sec:evalsetting}).

We evaluate if these results are as pronounced as the observations based on \Cref{tab:mdist} as described above.
Overall, the results of the placebo prompting (\Pprompts\ prompts) in comparison to \Nprompts\ prompts do not reveal any notable differences for specific attribute values (see \Cref{sec:apppla}). 
The scores remain consistently stable across these comparisons.
Consequently, we conclude that the changes in model behavior are indeed more systematic for prompting with specific socio-demographic attributes than when irrelevant additional information is included.

The changes resulting from placebo prompting in comparison to \Nprompts\ prompts appear to be substantial, with some discrepancies even surpassing those related to socio-demographic prompting.
This pattern is more pronounced within the offensiveness task.
The discrepancies arise from the analysis that focuses on the absolute values of the differences, regardless of their direction, thus capturing notable fluctuations that may not reflect a consistent trend.

\subsection{RQ3: \RQthreeText}
We investigate how task properties influence the role of demographic information in prompting and whether patterns remain consistent across the tasks.
Our general conclusion indicates that there is no clearly distinguished default persona of the models for both tasks.
\Cref{tab:def} shows that results are in general consistent across the two tasks.
However, the analysis regarding RQ1 also highlights some task-specific tendencies.
Notably, there are also differences in the prediction distances when comparing demographic prompting to no-information prompting across the two tasks.
The average prediction difference associated with socio-demographic prompting of GPT-4o for offensiveness rating is calculated to be 0.19.
In contrast, the average for politeness rating is substantially higher at 0.25.
Similar results are evident for the model Claude (0.16 for offensiveness rating and 0.21 for politeness rating).
This suggests that, in general, these LLMs are more influenced by demographic information at rating politeness than rating offensiveness.

\subsection{RQ4: \RQfourText}
The patterns identified in the analysis presented in \Cref{tab:def} demonstrate consistency across the two models examined, particularly in their ability to replicate human annotations.
Notably, \Cref{tab:mdist} highlights specific behaviors of the LLMs in response to socio-demographic prompting.
However, each of these distinct behaviors is only observed in one of the models.
Overall, both tested LLMs do not display clear default personas.
Thus, the models remain consistent in this aspect.

\section{Conclusion and Future Work}
\label{sec:conclusion}
In this paper, we present an analysis of the effect of socio-demographic prompting on tasks in the \textsc{Popquorn} data set.
Our findings show that demographic prompting exerts measurable effects on the annotation behaviors of large language models.
We contrast this to placebo prompting, which elicits no consistent changes across various attributes.
Specifically, our analyses reveal that LLMs show variations in annotation based on demographic attributes, particularly for gender, race, and age.
While we cannot infer one concrete, unique default persona, we conclude that large language models do not represent all members of a society alike, and that socio-demographic prompting does influence the result in a structured manner.
This stands in contrast to the results from previous studies, such as those by \citet{beck-etal-2024-sensitivity} and \citet{mukherjee2024culturalconditioningplaceboeffectiveness}, which report no consistent patterns.

Furthermore, our results echo some of the findings from \citet{sun-etal-2025-sociodemographic} regarding gender influences and that predictions tend to align more closely with the perceptions of White individuals.
However, our analysis includes a broader range of demographic attributes.
\citet{hu-collier-2024-quantifying} observe that while persona variables produce modest improvements, their limited explanatory power aligns with our findings on demographic prompting.
Similarly, \citet{movva-etal-2024-annotation} highlight challenges in aligning LLMs with safety evaluations across demographics, underscoring the need to understand these impacts on model behavior.
Together, these insights confirm the necessity of exploring demographic influences in LLM outputs.

Our study highlights biases in the annotation behavior of LLMs regarding demographics.
These biases raise critical concerns about the perpetuation of racial inequities in applications of these models.
The discrepancies we observe regarding different demographic groups suggest that LLMs may entrench existing biases rather than mitigate them.
This reinforces societal norms that marginalize diversity, demonstrating that models struggle to accurately represent certain demographic groups.

Not all our findings are consistent across models.
While this could be considered an issue that is inherent to LLMs, it also presents itself with a substantial challenge: depending on the model that is used by an end-user, the impact of socio-demographic information varies, and different models default to different demographic information.

Consequently, we advocate for future research to concentrate on developing reliable tools to measure the influence of both implicitly and explicitly provided demographic information.
Making demographic information explicit and ensuring its accurate interpretation by models is crucial not only for addressing hidden biases but also for creating models that genuinely reflect the diversity and variance among individuals.
Ultimately, our work underscores the necessity for bias-aware design in training models to provide equitable representation across demographic groups and to mitigate the risks of reinforcing existing societal inequities.

\section*{Limitations}
Our study considers only a limited set of models and
variables. Particularly limiting is the set of attributes available in the
data set, which are also culturally biased towards the USA.
Another considerable
limitation is the unequal representation of demographic subgroups.
We additionally only analyze a sample of the human raters to represent the text instances equally, which might lead to a skewed representation of certain demographics. However, our sample maintains a certain degree of representativeness of the demographics in  \textsc{Popquorn}. Using a larger sample would increase robustness of our results.

While the decision to further analyze only demographics with over 100 annotated instances aimed to enhance the manageability and interpretability of our results, it may introduce concerns regarding the statistical confidence of findings, particularly for the non-binary group, which marginally meets this threshold.
Fluctuations in the observed effects may also partially arise from the group's under-representation in this evaluation.

A limitation of our analysis is the lack of post-hoc statistical testing, which could strengthen our findings.
Although such tests are often omitted in regression model comparisons with few predictors, examining them could analyze whether assumptions like homoscedasticity and collinearity might have been violated.
It is unlikely that the results or the conclusion of our work is influenced, however, this can be a possible factor that might have undermined the overall process.

Our experiments are conducted with only two large language models and each experiment was run just once, for pragmatic reasons stemming from limited financial resources.
A large scale analysis across larger set of tasks would mitigate this issue.

Finally, the reported differences in
annotation outputs are relatively small compared to the overall Likert
scale range of 1--5.  This raises questions about the practical
significance of some of the observed effects, necessitating further
exploration to understand their impact in real-world applications.

\section*{Ethical Considerations}
Our work has the goal to make challenges transparent which the use of
large language models has. A considerable limitation from an ethical
perspective is that the variables we consider are not relevant across
cultures in the world. The response options of the ``race'' variable, in particular, are specific to US American society. Other variables might be
relevant in other cultures in the world that we are not aware of. We
therefore suggest to conduct future studies that consider more open
sets of variables that describe the diversity of users of language model-based
systems.

Beyond the limitations of variable selection, our findings also have implications for issues of inequality in the context of large language models.
If models effectively mimic particular demographic groups, this can lead to the privileging of the viewpoints and experiences of those groups, potentially marginalizing individuals who are not adequately represented.
Furthermore, if the responses are influenced by the inclusion of certain demographic information over others, it suggests that some demographic categories are represented as more ``normal'' or typical than others.
This pattern mirrors historical inequities that persist in new technologies, revealing an unsettling continuity in the ways that cultural biases may be perpetuated. 

ChatGPT \citep{chatgpt} was used to gain inspiration for formulations of our initial notes for the text of some sections of this paper, as well as to find typos.

\section*{Acknowledgments}
This work has been supported by the German Research Foundation (DFG) in the projects KL2869/1--2 (CEAT, project number 380093645), KL2869/5--1 (FIBISS, 438135827), KL2869/11--1 (ITEM, 513384754), KL2869/12--1 (EMCONA, 516512112), KL2869/13--1 (INPROMPT, 521755488).

\bibliography{lit}

\begin{thebibliography}{67}
\providecommand{\natexlab}[1]{#1}

\bibitem[{Abercrombie et~al.(2024)Abercrombie, Basile, Bernadi, Dudy, Frenda,
  Havens, and Tonelli}]{nlperspectives-2024-perspectivist}
Gavin Abercrombie, Valerio Basile, Davide Bernadi, Shiran Dudy, Simona Frenda,
  Lucy Havens, and Sara Tonelli, editors. 2024.
\newblock \href {https://aclanthology.org/2024.nlperspectives-1.0}
  {\emph{Proceedings of the 3rd Workshop on Perspectivist Approaches to NLP
  (NLPerspectives) @ LREC-COLING 2024}}. ELRA and ICCL, Torino, Italia.

\bibitem[{Anthropic(2024)}]{claude}
Anthropic. 2024.
\newblock \href {https://www.anthropic.com/news/claude-3-5-sonnet} {{Claude 3.5
  Sonnet}}.
\newblock Large language model.

\bibitem[{Baan et~al.(2024)Baan, Fern{\'a}ndez, Plank, and
  Aziz}]{baan-etal-2024-interpreting}
Joris Baan, Raquel Fern{\'a}ndez, Barbara Plank, and Wilker Aziz. 2024.
\newblock \href {https://aclanthology.org/2024.eacl-short.24} {Interpreting
  predictive probabilities: Model confidence or human label variation?}
\newblock In \emph{Proceedings of the 18th Conference of the European Chapter
  of the Association for Computational Linguistics (Volume 2: Short Papers)},
  pages 268--277, St. Julian{'}s, Malta. Association for Computational
  Linguistics.

\bibitem[{Bagdon et~al.(2024)Bagdon, Karmalkar, Gurulingappa, and
  Klinger}]{bagdon-etal-2024-expert}
Christopher Bagdon, Prathamesh Karmalkar, Harsha Gurulingappa, and Roman
  Klinger. 2024.
\newblock \href {https://doi.org/10.18653/v1/2024.naacl-long.439} {{``}you are
  an expert annotator{''}: Automatic best{--}worst-scaling annotations for
  emotion intensity modeling}.
\newblock In \emph{Proceedings of the 2024 Conference of the North American
  Chapter of the Association for Computational Linguistics: Human Language
  Technologies (Volume 1: Long Papers)}, pages 7924--7936, Mexico City, Mexico.
  Association for Computational Linguistics.

\bibitem[{Banerjee et~al.(2022)Banerjee, Bhattacharya, and Bera}]{Banerjee2022}
Abhishek Banerjee, Uttaran Bhattacharya, and Aniket Bera. 2022.
\newblock \href
  {https://aaai.org/papers/00003-learning-unseen-emotions-from-gestures-via-semantically-conditioned-zero-shot-perception-with-adversarial-autoencoders/}
  {Learning unseen emotions from gestures via semantically-conditioned
  zero-shot perception with adversarial autoencoders}.
\newblock In \emph{Proceedings of the AAAI Conference on Artificial
  Intelligence, 36}.

\bibitem[{Barei\ss{} et~al.(2024)Barei\ss{}, Klinger, and Barnes}]{Bareiss2024}
Patrick Barei\ss{}, Roman Klinger, and Jeremy Barnes. 2024.
\newblock \href {https://doi.org/10.1145/3589335.3651902} {English prompts are
  better for nli-based zero-shot emotion classification than target-language
  prompts}.
\newblock In \emph{Companion Proceedings of the ACM on Web Conference 2024},
  WWW '24, page 1318–1326, New York, NY, USA. Association for Computing
  Machinery.

\bibitem[{Baumler et~al.(2023)Baumler, Sotnikova, and
  Daum{\'e}~III}]{baumler-etal-2023-examples}
Connor Baumler, Anna Sotnikova, and Hal Daum{\'e}~III. 2023.
\newblock \href {https://doi.org/10.18653/v1/2023.findings-acl.658} {Which
  examples should be multiply annotated? active learning when annotators may
  disagree}.
\newblock In \emph{Findings of the Association for Computational Linguistics:
  ACL 2023}, pages 10352--10371, Toronto, Canada. Association for Computational
  Linguistics.

\bibitem[{Beck et~al.(2024)Beck, Schuff, Lauscher, and
  Gurevych}]{beck-etal-2024-sensitivity}
Tilman Beck, Hendrik Schuff, Anne Lauscher, and Iryna Gurevych. 2024.
\newblock \href {https://aclanthology.org/2024.eacl-long.159} {Sensitivity,
  performance, robustness: Deconstructing the effect of sociodemographic
  prompting}.
\newblock In \emph{Proceedings of the 18th Conference of the European Chapter
  of the Association for Computational Linguistics (Volume 1: Long Papers)},
  pages 2589--2615, St. Julian{'}s, Malta. Association for Computational
  Linguistics.

\bibitem[{Beurer-Kellner et~al.(2023)Beurer-Kellner, Fischer, and
  Vechev}]{Beurer2024}
Luca Beurer-Kellner, Marc Fischer, and Martin Vechev. 2023.
\newblock \href {https://doi.org/10.1145/3591300} {Prompting is programming: A
  query language for large language models}.
\newblock \emph{Proceedings of the ACM on Programming Languages}, 7(Issue
  PLDI):1946--1969.

\bibitem[{Bizzoni et~al.(2022)Bizzoni, Lassen, Peura, Thomsen, and
  Nielbo}]{bizzoni-etal-2022-predicting}
Yuri Bizzoni, Ida~Marie Lassen, Telma Peura, Mads~Rosendahl Thomsen, and
  Kristoffer Nielbo. 2022.
\newblock \href {https://aclanthology.org/2022.nlperspectives-1.3} {Predicting
  literary quality how perspectivist should we be?}
\newblock In \emph{Proceedings of the 1st Workshop on Perspectivist Approaches
  to NLP @LREC2022}, pages 20--25, Marseille, France. European Language
  Resources Association.

\bibitem[{Brown et~al.(2020)Brown, Mann, Ryder, Subbiah, Kaplan, Dhariwal,
  Neelakantan, Shyam, Sastry, Askell, Agarwal, Herbert-Voss, Krueger, Henighan,
  Child, Ramesh, Ziegler, Wu, Winter, Hesse, Chen, Sigler, Litwin, Gray, Chess,
  Clark, Berner, McCandlish, Radford, Sutskever, and
  Amodei}]{brown2020languagemodelsfewshotlearners}
Tom Brown, Benjamin Mann, Nick Ryder, Melanie Subbiah, Jared~D Kaplan, Prafulla
  Dhariwal, Arvind Neelakantan, Pranav Shyam, Girish Sastry, Amanda Askell,
  Sandhini Agarwal, Ariel Herbert-Voss, Gretchen Krueger, Tom Henighan, Rewon
  Child, Aditya Ramesh, Daniel Ziegler, Jeffrey Wu, Clemens Winter, Chris
  Hesse, Mark Chen, Eric Sigler, Mateusz Litwin, Scott Gray, Benjamin Chess,
  Jack Clark, Christopher Berner, Sam McCandlish, Alec Radford, Ilya Sutskever,
  and Dario Amodei. 2020.
\newblock \href
  {https://proceedings.neurips.cc/paper_files/paper/2020/file/1457c0d6bfcb4967418bfb8ac142f64a-Paper.pdf}
  {Language models are few-shot learners}.
\newblock In \emph{Advances in Neural Information Processing Systems},
  volume~33, pages 1877--1901. Curran Associates, Inc.

\bibitem[{Ceron et~al.(2024)Ceron, Falk, Barić, Nikolaev, and
  Padó}]{ceron2024promptbrittlenessevaluatingreliability}
Tanise Ceron, Neele Falk, Ana Barić, Dmitry Nikolaev, and Sebastian Padó.
  2024.
\newblock \href {https://arxiv.org/abs/2402.17649} {Beyond prompt brittleness:
  Evaluating the reliability and consistency of political worldviews in llms}.
\newblock \emph{Preprint}, arXiv:2402.17649.

\bibitem[{Chen and Shu(2023)}]{chen-shu-2023-promptda}
Canyu Chen and Kai Shu. 2023.
\newblock \href {https://doi.org/10.18653/v1/2023.eacl-main.41} {{P}rompt{DA}:
  Label-guided data augmentation for prompt-based few shot learners}.
\newblock In \emph{Proceedings of the 17th Conference of the European Chapter
  of the Association for Computational Linguistics}, pages 562--574, Dubrovnik,
  Croatia. Association for Computational Linguistics.

\bibitem[{Cheng et~al.(2023)Cheng, Durmus, and
  Jurafsky}]{cheng-etal-2023-marked}
Myra Cheng, Esin Durmus, and Dan Jurafsky. 2023.
\newblock \href {https://doi.org/10.18653/v1/2023.acl-long.84} {Marked
  personas: Using natural language prompts to measure stereotypes in language
  models}.
\newblock In \emph{Proceedings of the 61st Annual Meeting of the Association
  for Computational Linguistics (Volume 1: Long Papers)}, pages 1504--1532,
  Toronto, Canada. Association for Computational Linguistics.

\bibitem[{Farquhar et~al.(2024)Farquhar, Kossen, Kuhn, and
  Gal}]{farquhar_detecting_2024}
Sebastian Farquhar, Jannik Kossen, Lorenz Kuhn, and Yarin Gal. 2024.
\newblock \href {https://doi.org/10.1038/s41586-024-07421-0} {Detecting
  hallucinations in large language models using semantic entropy}.
\newblock \emph{Nature}, 630:625--630.

\bibitem[{Fei et~al.(2023)Fei, Li, Liu, Bing, Li, and
  Chua}]{fei-etal-2023-reasoning}
Hao Fei, Bobo Li, Qian Liu, Lidong Bing, Fei Li, and Tat-Seng Chua. 2023.
\newblock \href {https://doi.org/10.18653/v1/2023.acl-short.101} {Reasoning
  implicit sentiment with chain-of-thought prompting}.
\newblock In \emph{Proceedings of the 61st Annual Meeting of the Association
  for Computational Linguistics (Volume 2: Short Papers)}, pages 1171--1182,
  Toronto, Canada. Association for Computational Linguistics.

\bibitem[{Feng et~al.(2023)Feng, Park, Liu, and
  Tsvetkov}]{feng-etal-2023-pretraining}
Shangbin Feng, Chan~Young Park, Yuhan Liu, and Yulia Tsvetkov. 2023.
\newblock \href {https://doi.org/10.18653/v1/2023.acl-long.656} {From
  pretraining data to language models to downstream tasks: Tracking the trails
  of political biases leading to unfair {NLP} models}.
\newblock In \emph{Proceedings of the 61st Annual Meeting of the Association
  for Computational Linguistics (Volume 1: Long Papers)}, pages 11737--11762,
  Toronto, Canada. Association for Computational Linguistics.

\bibitem[{Finn and Louviere(1992)}]{Louv-Food-BWS}
Adam Finn and Jordan~J. Louviere. 1992.
\newblock \href {https://doi.org/10.1177/074391569201100202} {Determining the
  appropriate response to evidence of public concern: The case of food safety}.
\newblock \emph{Journal of Public Policy \& Marketing}, 11(2):12--25.

\bibitem[{Fleisig et~al.(2024)Fleisig, Blodgett, Klein, and
  Talat}]{fleisig-etal-2024-perspectivist}
Eve Fleisig, Su~Lin Blodgett, Dan Klein, and Zeerak Talat. 2024.
\newblock \href {https://doi.org/10.18653/v1/2024.naacl-long.126} {The
  perspectivist paradigm shift: Assumptions and challenges of capturing human
  labels}.
\newblock In \emph{Proceedings of the 2024 Conference of the North American
  Chapter of the Association for Computational Linguistics: Human Language
  Technologies (Volume 1: Long Papers)}, pages 2279--2292, Mexico City, Mexico.
  Association for Computational Linguistics.

\bibitem[{Frenda et~al.(2023)Frenda, Pedrani, Basile, Lo, Cignarella, Panizzon,
  Marco, Scarlini, Patti, Bosco, and Bernardi}]{frenda-etal-2023-epic}
Simona Frenda, Alessandro Pedrani, Valerio Basile, Soda~Marem Lo,
  Alessandra~Teresa Cignarella, Raffaella Panizzon, Cristina Marco, Bianca
  Scarlini, Viviana Patti, Cristina Bosco, and Davide Bernardi. 2023.
\newblock \href {https://doi.org/10.18653/v1/2023.acl-long.774} {{EPIC}:
  Multi-perspective annotation of a corpus of irony}.
\newblock In \emph{Proceedings of the 61st Annual Meeting of the Association
  for Computational Linguistics (Volume 1: Long Papers)}, pages 13844--13857,
  Toronto, Canada. Association for Computational Linguistics.

\bibitem[{Gruber et~al.(2024)Gruber, Hechinger, Assenmacher, Kauermann, and
  Plank}]{gruber-etal-2024-labels}
Cornelia Gruber, Katharina Hechinger, Matthias Assenmacher, G{\"o}ran
  Kauermann, and Barbara Plank. 2024.
\newblock \href {https://aclanthology.org/2024.unimplicit-1.2} {More labels or
  cases? assessing label variation in natural language inference}.
\newblock In \emph{Proceedings of the Third Workshop on Understanding Implicit
  and Underspecified Language}, pages 22--32, Malta. Association for
  Computational Linguistics.

\bibitem[{Hada et~al.(2021)Hada, Sudhir, Mishra, Yannakoudakis, Mohammad, and
  Shutova}]{hada-etal-2021-ruddit}
Rishav Hada, Sohi Sudhir, Pushkar Mishra, Helen Yannakoudakis, Saif~M.
  Mohammad, and Ekaterina Shutova. 2021.
\newblock \href {https://doi.org/10.18653/v1/2021.acl-long.210} {Ruddit:
  {N}orms of offensiveness for {E}nglish {R}eddit comments}.
\newblock In \emph{Proceedings of the 59th Annual Meeting of the Association
  for Computational Linguistics and the 11th International Joint Conference on
  Natural Language Processing (Volume 1: Long Papers)}, pages 2700--2717,
  Online. Association for Computational Linguistics.

\bibitem[{Hovy et~al.(2013)Hovy, Berg-Kirkpatrick, Vaswani, and
  Hovy}]{hovy-etal-2013-learning}
Dirk Hovy, Taylor Berg-Kirkpatrick, Ashish Vaswani, and Eduard Hovy. 2013.
\newblock \href {https://aclanthology.org/N13-1132} {Learning whom to trust
  with {MACE}}.
\newblock In \emph{Proceedings of the 2013 Conference of the North {A}merican
  Chapter of the Association for Computational Linguistics: Human Language
  Technologies}, pages 1120--1130, Atlanta, Georgia. Association for
  Computational Linguistics.

\bibitem[{Hu and Collier(2024)}]{hu-collier-2024-quantifying}
Tiancheng Hu and Nigel Collier. 2024.
\newblock \href {https://doi.org/10.18653/v1/2024.acl-long.554} {Quantifying
  the persona effect in {LLM} simulations}.
\newblock In \emph{Proceedings of the 62nd Annual Meeting of the Association
  for Computational Linguistics (Volume 1: Long Papers)}, pages 10289--10307,
  Bangkok, Thailand. Association for Computational Linguistics.

\bibitem[{Johfre and Freese(2021)}]{doi:10.1177/0081175020982632}
Sasha~Shen Johfre and Jeremy Freese. 2021.
\newblock \href {https://doi.org/10.1177/0081175020982632} {Reconsidering the
  reference category}.
\newblock \emph{Sociological Methodology}, 51(2):253--269.

\bibitem[{Klinger(2023)}]{klinger-2023-event}
Roman Klinger. 2023.
\newblock \href {https://doi.org/10.18653/v1/2023.bigpicture-1.1} {Where are we
  in event-centric emotion analysis? bridging emotion role labeling and
  appraisal-based approaches}.
\newblock In \emph{Proceedings of the Big Picture Workshop}, pages 1--17,
  Singapore. Association for Computational Linguistics.

\bibitem[{Lee et~al.(2023)Lee, An, and Thorne}]{lee-etal-2023-large}
Noah Lee, Na~Min An, and James Thorne. 2023.
\newblock \href {https://doi.org/10.18653/v1/2023.emnlp-main.278} {Can large
  language models capture dissenting human voices?}
\newblock In \emph{Proceedings of the 2023 Conference on Empirical Methods in
  Natural Language Processing}, pages 4569--4585, Singapore. Association for
  Computational Linguistics.

\bibitem[{Li(2023)}]{li-2023-practical}
Yinheng Li. 2023.
\newblock \href {https://aclanthology.org/2023.ranlp-1.69} {A practical survey
  on zero-shot prompt design for in-context learning}.
\newblock In \emph{Proceedings of the 14th International Conference on Recent
  Advances in Natural Language Processing}, pages 641--647, Varna, Bulgaria.
  INCOMA Ltd., Shoumen, Bulgaria.

\bibitem[{Liu(2012)}]{Liu2012}
Bing Liu. 2012.
\newblock \href {https://doi.org/10.1017/cbo9781139084789} {\emph{Sentiment
  Analysis: Mining Opinions, Sentiments, and Emotions}}.
\newblock Cambridge University Press.

\bibitem[{Liu et~al.(2023)Liu, Yuan, Fu, Jiang, Hayashi, and
  Neubig}]{liu2021pretrainpromptpredictsystematic}
Pengfei Liu, Weizhe Yuan, Jinlan Fu, Zhengbao Jiang, Hiroaki Hayashi, and
  Graham Neubig. 2023.
\newblock \href {https://doi.org/10.1145/3560815} {Pre-train, prompt, and
  predict: A systematic survey of prompting methods in natural language
  processing}.
\newblock \emph{ACM Computing Surveys}, 55(9).

\bibitem[{Louviere et~al.(2015)Louviere, Flyn, and Marley}]{Louv-BWS}
Jordan~J. Louviere, Terry~N. Flyn, and A.A.J. Marley. 2015.
\newblock \href
  {http://www.cambridge.org/us/academic/subjects/economics/econometrics-statistics-and-mathematical-economics/best-worst-scaling-theory-methods-and-applications}
  {\emph{Best-Worst Scaling -- Theory, Methods and Applications}}.
\newblock Cambridge University Press.

\bibitem[{Ma et~al.(2022)Ma, Wang, Cao, Li, Chen, Wang, and
  Shao}]{ma-etal-2022-prompt}
Yubo Ma, Zehao Wang, Yixin Cao, Mukai Li, Meiqi Chen, Kun Wang, and Jing Shao.
  2022.
\newblock \href {https://doi.org/10.18653/v1/2022.acl-long.466} {{P}rompt for
  extraction? {PAIE}: {P}rompting argument interaction for event argument
  extraction}.
\newblock In \emph{Proceedings of the 60th Annual Meeting of the Association
  for Computational Linguistics (Volume 1: Long Papers)}, pages 6759--6774,
  Dublin, Ireland. Association for Computational Linguistics.

\bibitem[{May et~al.(2024)May, Flek, and Welch}]{may-etal-2024-perspectivist}
Marlon May, Lucie Flek, and Charles Welch. 2024.
\newblock \href {https://aclanthology.org/2024.nlperspectives-1.4} {A
  perspectivist corpus of numbers in social judgements}.
\newblock In \emph{Proceedings of the 3rd Workshop on Perspectivist Approaches
  to NLP (NLPerspectives) @ LREC-COLING 2024}, pages 42--48, Torino, Italia.
  ELRA and ICCL.

\bibitem[{Motoki et~al.(2023)Motoki, Pinho~Neto, and Rodrigues}]{Motoki2023}
Fabio Motoki, Valdemar Pinho~Neto, and Victor Rodrigues. 2023.
\newblock \href {https://doi.org/10.1007/s11127-023-01097-2} {More human than
  human: measuring chatgpt political bias}.
\newblock \emph{Public Choice}, 198:3–23.

\bibitem[{Movva et~al.(2024)Movva, Koh, and
  Pierson}]{movva-etal-2024-annotation}
Rajiv Movva, Pang~Wei Koh, and Emma Pierson. 2024.
\newblock \href {https://doi.org/10.18653/v1/2024.emnlp-main.511} {Annotation
  alignment: Comparing {LLM} and human annotations of conversational safety}.
\newblock In \emph{Proceedings of the 2024 Conference on Empirical Methods in
  Natural Language Processing}, pages 9048--9062, Miami, Florida, USA.
  Association for Computational Linguistics.

\bibitem[{Mukherjee et~al.(2024)Mukherjee, Adilazuarda, Sitaram, Bali, Aji, and
  Choudhury}]{mukherjee2024culturalconditioningplaceboeffectiveness}
Sagnik Mukherjee, Muhammad~Farid Adilazuarda, Sunayana Sitaram, Kalika Bali,
  Alham~Fikri Aji, and Monojit Choudhury. 2024.
\newblock \href {https://arxiv.org/abs/2406.11661} {Cultural conditioning or
  placebo? on the effectiveness of socio-demographic prompting}.
\newblock \emph{Preprint}, arXiv:2406.11661.

\bibitem[{Muscato et~al.(2024)Muscato, Mala, Marchiori~Manerba, Gezici, and
  Giannotti}]{muscato-etal-2024-overview}
Benedetta Muscato, Chandana~Sree Mala, Marta Marchiori~Manerba, Gizem Gezici,
  and Fosca Giannotti. 2024.
\newblock \href {https://aclanthology.org/2024.nlperspectives-1.5} {An overview
  of recent approaches to enable diversity in large language models through
  aligning with human perspectives}.
\newblock In \emph{Proceedings of the 3rd Workshop on Perspectivist Approaches
  to NLP (NLPerspectives) @ LREC-COLING 2024}, pages 49--55, Torino, Italia.
  ELRA and ICCL.

\bibitem[{Neuman(2015)}]{neuman-2015-personality}
Yair Neuman. 2015.
\newblock \href {https://aclanthology.org/D15-2002} {Personality research for
  {NLP}}.
\newblock In \emph{Proceedings of the 2015 Conference on Empirical Methods in
  Natural Language Processing: Tutorial Abstracts}, Lisbon, Portugal.
  Association for Computational Linguistics.

\bibitem[{OpenAI(2024{\natexlab{a}})}]{chatgpt}
OpenAI. 2024{\natexlab{a}}.
\newblock \href {https://chat.openai.com/chat} {{ChatGPT (model GPT4o-mini)}}.
\newblock Large language model.

\bibitem[{OpenAI(2024{\natexlab{b}})}]{gpt4o}
OpenAI. 2024{\natexlab{b}}.
\newblock \href {https://openai.com/index/hello-gpt-4o/} {{GPT-4o}}.
\newblock Large language model.

\bibitem[{Palatucci et~al.(2009)Palatucci, Pomerleau, Hinton, and
  Mitchell}]{NIPS2009_1543843a}
Mark Palatucci, Dean Pomerleau, Geoffrey~E Hinton, and Tom~M Mitchell. 2009.
\newblock \href
  {https://proceedings.neurips.cc/paper_files/paper/2009/file/1543843a4723ed2ab08e18053ae6dc5b-Paper.pdf}
  {Zero-shot learning with semantic output codes}.
\newblock In \emph{Advances in Neural Information Processing Systems},
  volume~22. Curran Associates, Inc.

\bibitem[{Paun et~al.(2018)Paun, Carpenter, Chamberlain, Hovy, Kruschwitz, and
  Poesio}]{paun-etal-2018-comparing}
Silviu Paun, Bob Carpenter, Jon Chamberlain, Dirk Hovy, Udo Kruschwitz, and
  Massimo Poesio. 2018.
\newblock \href {https://doi.org/10.1162/tacl_a_00040} {Comparing {B}ayesian
  models of annotation}.
\newblock \emph{Transactions of the Association for Computational Linguistics},
  6:571--585.

\bibitem[{Pei and Jurgens(2023)}]{pei-jurgens-2023-annotator}
Jiaxin Pei and David Jurgens. 2023.
\newblock \href {https://doi.org/10.18653/v1/2023.law-1.25} {When do annotator
  demographics matter? measuring the influence of annotator demographics with
  the {POPQUORN} dataset}.
\newblock In \emph{Proceedings of the 17th Linguistic Annotation Workshop
  (LAW-XVII)}, pages 252--265, Toronto, Canada. Association for Computational
  Linguistics.

\bibitem[{Peng et~al.(2024)Peng, Sun, Loftus, and
  Plank}]{peng-etal-2024-different}
Siyao Peng, Zihang Sun, Sebastian Loftus, and Barbara Plank. 2024.
\newblock \href {https://aclanthology.org/2024.unimplicit-1.7} {Different
  tastes of entities: Investigating human label variation in named entity
  annotations}.
\newblock In \emph{Proceedings of the Third Workshop on Understanding Implicit
  and Underspecified Language}, pages 73--81, Malta. Association for
  Computational Linguistics.

\bibitem[{Petroni et~al.(2019)Petroni, Rockt{\"a}schel, Riedel, Lewis, Bakhtin,
  Wu, and Miller}]{petroni-etal-2019-language}
Fabio Petroni, Tim Rockt{\"a}schel, Sebastian Riedel, Patrick Lewis, Anton
  Bakhtin, Yuxiang Wu, and Alexander Miller. 2019.
\newblock \href {https://doi.org/10.18653/v1/D19-1250} {Language models as
  knowledge bases?}
\newblock In \emph{Proceedings of the 2019 Conference on Empirical Methods in
  Natural Language Processing and the 9th International Joint Conference on
  Natural Language Processing (EMNLP-IJCNLP)}, pages 2463--2473, Hong Kong,
  China. Association for Computational Linguistics.

\bibitem[{Plank(2022)}]{plank-2022-problem}
Barbara Plank. 2022.
\newblock \href {https://doi.org/10.18653/v1/2022.emnlp-main.731} {The
  {``}problem{''} of human label variation: On ground truth in data, modeling
  and evaluation}.
\newblock In \emph{Proceedings of the 2022 Conference on Empirical Methods in
  Natural Language Processing}, pages 10671--10682, Abu Dhabi, United Arab
  Emirates. Association for Computational Linguistics.

\bibitem[{Plaza-del Arco et~al.(2024)Plaza-del Arco, Cercas~Curry,
  Cercas~Curry, and Hovy}]{plaza-del-arco-etal-2024-emotion}
Flor~Miriam Plaza-del Arco, Alba~A. Cercas~Curry, Amanda Cercas~Curry, and Dirk
  Hovy. 2024.
\newblock \href {https://aclanthology.org/2024.lrec-main.506} {Emotion analysis
  in {NLP}: Trends, gaps and roadmap for future directions}.
\newblock In \emph{Proceedings of the 2024 Joint International Conference on
  Computational Linguistics, Language Resources and Evaluation (LREC-COLING
  2024)}, pages 5696--5710, Torino, Italia. ELRA and ICCL.

\bibitem[{Plaza-del Arco et~al.(2022)Plaza-del Arco, Mart{\'\i}n-Valdivia, and
  Klinger}]{plaza-del-arco-etal-2022-natural}
Flor~Miriam Plaza-del Arco, Mar{\'\i}a-Teresa Mart{\'\i}n-Valdivia, and Roman
  Klinger. 2022.
\newblock \href {https://aclanthology.org/2022.coling-1.592} {Natural language
  inference prompts for zero-shot emotion classification in text across
  corpora}.
\newblock In \emph{Proceedings of the 29th International Conference on
  Computational Linguistics}, pages 6805--6817, Gyeongju, Republic of Korea.
  International Committee on Computational Linguistics.

\bibitem[{Plepi et~al.(2022)Plepi, Neuendorf, Flek, and
  Welch}]{plepi-etal-2022-unifying}
Joan Plepi, B{\'e}la Neuendorf, Lucie Flek, and Charles Welch. 2022.
\newblock \href {https://doi.org/10.18653/v1/2022.emnlp-main.500} {Unifying
  data perspectivism and personalization: An application to social norms}.
\newblock In \emph{Proceedings of the 2022 Conference on Empirical Methods in
  Natural Language Processing}, pages 7391--7402, Abu Dhabi, United Arab
  Emirates. Association for Computational Linguistics.

\bibitem[{Romberg(2022)}]{romberg-2022-perspective}
Julia Romberg. 2022.
\newblock \href {https://aclanthology.org/2022.argmining-1.11} {Is your
  perspective also my perspective? enriching prediction with subjectivity}.
\newblock In \emph{Proceedings of the 9th Workshop on Argument Mining}, pages
  115--125, Online and in Gyeongju, Republic of Korea. International Conference
  on Computational Linguistics.

\bibitem[{Sachdeva et~al.(2022)Sachdeva, Barreto, Bacon, Sahn, von Vacano, and
  Kennedy}]{sachdeva-etal-2022-measuring}
Pratik Sachdeva, Renata Barreto, Geoff Bacon, Alexander Sahn, Claudia von
  Vacano, and Chris Kennedy. 2022.
\newblock \href {https://aclanthology.org/2022.nlperspectives-1.11} {The
  measuring hate speech corpus: Leveraging rasch measurement theory for data
  perspectivism}.
\newblock In \emph{Proceedings of the 1st Workshop on Perspectivist Approaches
  to NLP @LREC2022}, pages 83--94, Marseille, France. European Language
  Resources Association.

\bibitem[{Santurkar et~al.(2023)Santurkar, Durmus, Ladhak, Lee, Liang, and
  Hashimoto}]{Santurkar2023}
Shibani Santurkar, Esin Durmus, Faisal Ladhak, Cinoo Lee, Percy Liang, and
  Tatsunori Hashimoto. 2023.
\newblock \href {https://proceedings.mlr.press/v202/santurkar23a.html} {Whose
  opinions do language models reflect?}
\newblock In \emph{Proceedings of the 40th International Conference on Machine
  Learning}, volume 202 of \emph{Proceedings of Machine Learning Research},
  pages 29971--30004. PMLR.

\bibitem[{Santy et~al.(2023)Santy, Liang, Le~Bras, Reinecke, and
  Sap}]{santy-etal-2023-nlpositionality}
Sebastin Santy, Jenny Liang, Ronan Le~Bras, Katharina Reinecke, and Maarten
  Sap. 2023.
\newblock \href {https://doi.org/10.18653/v1/2023.acl-long.505}
  {{NLP}ositionality: Characterizing design biases of datasets and models}.
\newblock In \emph{Proceedings of the 61st Annual Meeting of the Association
  for Computational Linguistics (Volume 1: Long Papers)}, pages 9080--9102,
  Toronto, Canada. Association for Computational Linguistics.

\bibitem[{Shen et~al.(2023)Shen, Tan, Wu, Zhang, Zhang, Xi, Lu, and
  Zhuang}]{shen-etal-2023-promptner}
Yongliang Shen, Zeqi Tan, Shuhui Wu, Wenqi Zhang, Rongsheng Zhang, Yadong Xi,
  Weiming Lu, and Yueting Zhuang. 2023.
\newblock \href {https://doi.org/10.18653/v1/2023.acl-long.698} {{P}rompt{NER}:
  Prompt locating and typing for named entity recognition}.
\newblock In \emph{Proceedings of the 61st Annual Meeting of the Association
  for Computational Linguistics (Volume 1: Long Papers)}, pages 12492--12507,
  Toronto, Canada. Association for Computational Linguistics.

\bibitem[{Shi et~al.(2020)Shi, Malioutov, and Irsoy}]{shi-etal-2020-semantic}
Tianze Shi, Igor Malioutov, and Ozan Irsoy. 2020.
\newblock \href {https://doi.org/10.18653/v1/2020.emnlp-main.610} {Semantic
  role labeling as syntactic dependency parsing}.
\newblock In \emph{Proceedings of the 2020 Conference on Empirical Methods in
  Natural Language Processing (EMNLP)}, pages 7551--7571, Online. Association
  for Computational Linguistics.

\bibitem[{Stubbs(2011)}]{stubbs-2011-mae}
Amber Stubbs. 2011.
\newblock \href {https://aclanthology.org/W11-0416} {{MAE} and {MAI}:
  Lightweight annotation and adjudication tools}.
\newblock In \emph{Proceedings of the 5th Linguistic Annotation Workshop},
  pages 129--133, Portland, Oregon, USA. Association for Computational
  Linguistics.

\bibitem[{Sun et~al.(2025)Sun, Pei, Choi, and
  Jurgens}]{sun-etal-2025-sociodemographic}
Huaman Sun, Jiaxin Pei, Minje Choi, and David Jurgens. 2025.
\newblock \href {https://aclanthology.org/2025.naacl-short.71/}
  {Sociodemographic prompting is not yet an effective approach for simulating
  subjective judgments with {LLM}s}.
\newblock In \emph{Proceedings of the 2025 Conference of the Nations of the
  Americas Chapter of the Association for Computational Linguistics: Human
  Language Technologies (Volume 2: Short Papers)}, pages 845--854, Albuquerque,
  New Mexico. Association for Computational Linguistics.

\bibitem[{Troiano et~al.(2023)Troiano, Oberl{\"a}nder, and
  Klinger}]{troiano-etal-2023-dimensional}
Enrica Troiano, Laura Oberl{\"a}nder, and Roman Klinger. 2023.
\newblock \href {https://doi.org/10.1162/coli_a_00461} {Dimensional modeling of
  emotions in text with appraisal theories: Corpus creation, annotation
  reliability, and prediction}.
\newblock \emph{Computational Linguistics}, 49(1):1--72.

\bibitem[{Troiano et~al.(2021)Troiano, Pad{\'o}, and
  Klinger}]{troiano-etal-2021-emotion}
Enrica Troiano, Sebastian Pad{\'o}, and Roman Klinger. 2021.
\newblock \href {https://aclanthology.org/2021.wassa-1.5} {Emotion ratings: How
  intensity, annotation confidence and agreements are entangled}.
\newblock In \emph{Proceedings of the Eleventh Workshop on Computational
  Approaches to Subjectivity, Sentiment and Social Media Analysis}, pages
  40--49, Online. Association for Computational Linguistics.

\bibitem[{Uma et~al.(2021)Uma, Fornaciari, Dumitrache, Miller, Chamberlain,
  Plank, Simpson, and Poesio}]{uma-etal-2021-semeval}
Alexandra Uma, Tommaso Fornaciari, Anca Dumitrache, Tristan Miller, Jon
  Chamberlain, Barbara Plank, Edwin Simpson, and Massimo Poesio. 2021.
\newblock \href {https://doi.org/10.18653/v1/2021.semeval-1.41}
  {{S}em{E}val-2021 task 12: Learning with disagreements}.
\newblock In \emph{Proceedings of the 15th International Workshop on Semantic
  Evaluation (SemEval-2021)}, pages 338--347, Online. Association for
  Computational Linguistics.

\bibitem[{Valette(2024)}]{valette-2024-perspectivism}
Mathieu Valette. 2024.
\newblock \href {https://aclanthology.org/2024.nlperspectives-1.12} {What does
  perspectivism mean? an ethical and methodological countercriticism}.
\newblock In \emph{Proceedings of the 3rd Workshop on Perspectivist Approaches
  to NLP (NLPerspectives) @ LREC-COLING 2024}, pages 111--115, Torino, Italia.
  ELRA and ICCL.

\bibitem[{von Oswald et~al.(2023)von Oswald, Niklasson, Randazzo, Sacramento,
  Mordvintsev, Zhmoginov, and
  Vladymyrov}]{vonoswald2023transformerslearnincontextgradient}
Johannes von Oswald, Eyvind Niklasson, Ettore Randazzo, Joao Sacramento,
  Alexander Mordvintsev, Andrey Zhmoginov, and Max Vladymyrov. 2023.
\newblock \href {https://proceedings.mlr.press/v202/von-oswald23a.html}
  {Transformers learn in-context by gradient descent}.
\newblock In \emph{Proceedings of the 40th International Conference on Machine
  Learning}, volume 202 of \emph{Proceedings of Machine Learning Research},
  pages 35151--35174. PMLR.

\bibitem[{Wei et~al.(2022)Wei, Tay, Bommasani, Raffel, Zoph, Borgeaud,
  Yogatama, Bosma, Zhou, Metzler, Chi, Hashimoto, Vinyals, Liang, Dean, and
  Fedus}]{wei2022emergent}
Jason Wei, Yi~Tay, Rishi Bommasani, Colin Raffel, Barret Zoph, Sebastian
  Borgeaud, Dani Yogatama, Maarten Bosma, Denny Zhou, Donald Metzler, Ed~H.
  Chi, Tatsunori Hashimoto, Oriol Vinyals, Percy Liang, Jeff Dean, and William
  Fedus. 2022.
\newblock \href {https://openreview.net/forum?id=yzkSU5zdwD} {Emergent
  abilities of large language models}.
\newblock \emph{Transactions on Machine Learning Research}.

\bibitem[{Wright et~al.(2024)Wright, Arora, Borenstein, Yadav, Belongie, and
  Augenstein}]{wright2024revealingfinegrainedvaluesopinions}
Dustin Wright, Arnav Arora, Nadav Borenstein, Srishti Yadav, Serge Belongie,
  and Isabelle Augenstein. 2024.
\newblock \href {https://arxiv.org/abs/2406.19238} {Revealing fine-grained
  values and opinions in large language models}.
\newblock \emph{Preprint}, arXiv:2406.19238.

\bibitem[{Xu et~al.(2023)Xu, T.y.s.s, Ichim, Risini, Plank, and
  Grabmair}]{xu-etal-2023-dissonance}
Shanshan Xu, Santosh T.y.s.s, Oana Ichim, Isabella Risini, Barbara Plank, and
  Matthias Grabmair. 2023.
\newblock \href {https://doi.org/10.18653/v1/2023.emnlp-main.594} {From
  dissonance to insights: Dissecting disagreements in rationale construction
  for case outcome classification}.
\newblock In \emph{Proceedings of the 2023 Conference on Empirical Methods in
  Natural Language Processing}, pages 9558--9576, Singapore. Association for
  Computational Linguistics.

\bibitem[{Yadav and Bethard(2018)}]{yadav-bethard-2018-survey}
Vikas Yadav and Steven Bethard. 2018.
\newblock \href {https://aclanthology.org/C18-1182} {A survey on recent
  advances in named entity recognition from deep learning models}.
\newblock In \emph{Proceedings of the 27th International Conference on
  Computational Linguistics}, pages 2145--2158, Santa Fe, New Mexico, USA.
  Association for Computational Linguistics.

\bibitem[{Yin et~al.(2019)Yin, Hay, and Roth}]{yin-etal-2019-benchmarking}
Wenpeng Yin, Jamaal Hay, and Dan Roth. 2019.
\newblock \href {https://doi.org/10.18653/v1/D19-1404} {Benchmarking zero-shot
  text classification: Datasets, evaluation and entailment approach}.
\newblock In \emph{Proceedings of the 2019 Conference on Empirical Methods in
  Natural Language Processing and the 9th International Joint Conference on
  Natural Language Processing (EMNLP-IJCNLP)}, pages 3914--3923, Hong Kong,
  China. Association for Computational Linguistics.

\end{thebibliography}

\appendix

\section{Additional Results}
\label{sec:appres}
In this section we provide additional details regarding the results of our experiments.

\subsection{Full Results for LLMs Mimicing Annotators}
\label{sec:appful}

\begin{table*}[t]
	\centering \small \setlength{\tabcolsep}{5pt}
	\renewcommand{\arraystretch}{1.1}
	\begin{tabularx}{\linewidth}{X ll ll}
		\toprule
		& \multicolumn{2}{c}{Offensiveness} & \multicolumn{2}{c}{Politeness}  \\
		\cmidrule(r){2-3}\cmidrule(l){4-5}
		Socio-Demographic Attribute & GPT-4 & Claude & GPT-4  & Claude  \\
		\cmidrule(r){1-1} \cmidrule(lr){2-2} \cmidrule(lr){3-3} \cmidrule(lr){4-4} \cmidrule(l){5-5} 
		\textbf{Intercept}                         & 0.45 (0.17)**  & 0.56 (0.19)** & 1.45 (0.19)*** & 1.53 (0.20)*** \\
		\textbf{Age (years)}                       & 0.01 (0.00)**  & 0.01 (0.00)** & 0.00 (0.00)    & 0.00 (0.00)    \\
		\textbf{Gender (ref: Male) }&&&& \\
		\hspace{1em}Female                            & 0.00 (0.04)    & -0.03 (0.04)  & -0.05 (0.03)   & -0.05 (0.03)   \\
		\hspace{1em}Non-binary                        & -0.06 (0.12)   & -0.01 (0.13)  & -0.05 (0.10)   & -0.05 (0.10)   \\
		\textbf{Race (ref: White)} &&&& \\
		\hspace{1em}Asian                             & 0.09 (0.08)    & 0.03 (0.08)   & -0.08 (0.07)   & 0.00 (0.07)    \\
		\hspace{1em}Black or African American         & 0.22 (0.06)*** & 0.19 (0.06)** & 0.14 (0.05)**  & 0.15 (0.05)**  \\
		\hspace{1em}Hispanic or Latino                & -0.11 (0.14)   & -0.05 (0.15)  & 0.09 (0.06)    & 0.12 (0.06)+   \\
		\hspace{1em}Other race/ethnicity              & -0.14 (0.13)   & -0.26 (0.14)+ & -0.17 (0.18)   & -0.15 (0.19)   \\
		\textbf{Occupation (ref: Employed)} &&&& \\
		\hspace{1em}Unemployed                        & 0.04 (0.07)    & -0.08 (0.07)  & -0.08 (0.05)   & -0.08 (0.06)   \\
		\hspace{1em}Homemaker                         & -0.07 (0.10)   & -0.02 (0.10)  & -0.03 (0.07)   & -0.06 (0.07)   \\
		\hspace{1em}Retired                           & -0.11 (0.07)   & -0.13 (0.08)  & 0.03 (0.06)    & 0.03 (0.06)    \\
		\hspace{1em}Self-employed                     & 0.04 (0.06)    & 0.02 (0.07)   & -0.03 (0.05)   & -0.03 (0.05)   \\
	    \hspace{1em}Student                           & 0.13 (0.08)    & 0.13 (0.09)   & -0.11 (0.08)   & -0.13 (0.08)   \\
		\hspace{1em}Other occupation                  & -0.05 (0.15)   & 0.02 (0.16)   & 0.08 (0.11)    & 0.11 (0.11)    \\
		\textbf{Education (ref: Less than high school)} &&&& \\
		\hspace{1em}High school diploma or equivalent & -0.01 (0.15)   & -0.08 (0.17)  & -0.34 (0.18)+  & -0.37 (0.19)*  \\
		\hspace{1em}College degree                    & 0.05 (0.15)    & -0.09 (0.17)  & -0.43 (0.18)*  & -0.48 (0.19)*  \\
		\hspace{1em}Graduate degree                   & 0.06 (0.16)    & -0.01 (0.17)  & -0.36 (0.18)*  & -0.44 (0.19)*  \\
		\hspace{1em}Other education                   & -0.02 (0.21)   & 0.01 (0.22)   & -0.50 (0.21)*  & -0.57 (0.22)** \\
		\cmidrule(){1-1}
		SD (instance intercepts)          & 0.44           & 0.42          & 0.37           & 0.33           \\
		SD (annotator intercepts)         & 0.23           & 0.27          & 0.30           & 0.32           \\
		Num. Obs.                          & 4500           & 4481          & 11151          & 11151          \\
		R2 Marg.                          & 0.014          & 0.016         & 0.012          & 0.014          \\
		R2 Cond.                          & 0.316          & 0.321         & 0.311          & 0.305          \\
		ICC                               & 0.3            & 0.3           & 0.3            & 0.3            \\
		\bottomrule
	\end{tabularx}
	\caption{Detailed analyses and coefficients indicating the effect of particular human
		demographic categories on the distance between human and LLM
		annotations. }
	\label{tab:deffull}
\end{table*}

\Cref{tab:deffull} provides additional statistics complementing those presented in \Cref{tab:def}.
Here, we describe these additional statistics in more detail.
At the bottom of \Cref{tab:deffull}, we present regression model fit statistics.
The numbers in parentheses represent the standard error of the estimates, which indicates the uncertainty associated with these estimates and is used for calculating statistical significance.

``Intercept'' reflects the overall intercept of the model, representing the expected distance between the LLM’s predictions and those of a human, assuming all reference categories and an average age of 0 with an average annotation skill on a prompt of average ambiguity.
The other coefficients serve as adjustments to this baseline value.
The statistical significance of the intercept itself is not meaningful, and it is the effect sizes that are of primary interest rather than the absolute values of the predicted distances.
A ``+'' sign indicates marginal significance (0.05 $\leq$ P $\leq$ 0.1). 
SD (Standard Deviation) refers to the standard deviations of the mixed-effect model’s random intercepts for instances and annotators.
The model assigns a unique intercept to each instance and annotator to account for uninteresting idiosyncrasies, with a mean of 0.
The SD value indicates the expected variability of LLM-human differences across instances and annotators.
``Num. Obs.'' denotes the number of rows (annotations) on which the regression model was conducted.
R2 Marg.\ and R2 Cond.\ are measures of the model's explanatory power.
R2 Marginal indicates how much variation is accounted for by the coefficients alone, which is very low, suggesting that these coefficients contribute little to predictive power despite some being statistically significant.
R2 Conditional, on the other hand, indicates the proportion of variance explained by both the coefficients and the random intercepts, which is significantly higher, demonstrating that the random intercepts contribute more substantially to the model's explanatory power.

The ICC (Intraclass Correlation Coefficient) indicates the extent to which the clustering (by instances and annotators) influences the outcomes (distances between human and LLM ratings).
It represents the ratio of total variance in the dependent variable attributed to variance between cluster means, as opposed to variance within clusters.
A value of 0.3 or higher suggests that the inclusion of random intercepts is necessary to account for clustering.
If the ICC were very low (less than 0.1), a simpler model could be justified.

\subsection{Results for Individual Prompts}
\label{sec:appind}
In examining the scores for the individual prompts presented in \Cref{tab:varoff} and \Cref{tab:varpol}, we observe that there are no notable differences between the two prompts.
This lack of variation in scores is a positive outcome, as it suggests consistency in the model's performance across different prompts.
Such uniformity reinforces the reliability of the results, indicating that the prompts used do not unduly influence the models' annotation behaviors.
This consistency allows us to have greater confidence in our findings and their implications regarding the impact of demographic information on model outputs.

\begin{table*}[t]
	\centering \small \setlength{\tabcolsep}{5pt}
	\renewcommand{\arraystretch}{1.1}
	\begin{tabularx}{\linewidth}{X ll ll}
		\toprule
		& \multicolumn{2}{c}{Offensiveness Prompt 1} & \multicolumn{2}{c}{Offensiveness Prompt 2}  \\
		\cmidrule(r){2-3}\cmidrule(l){4-5}
		Socio-Demographic Attribute & GPT-4 & Claude & GPT-4  & Claude  \\
		\cmidrule(r){1-1} \cmidrule(lr){2-2} \cmidrule(lr){3-3} \cmidrule(lr){4-4} \cmidrule(l){5-5} 
		\textbf{Intercept}                         & 0.51 (0.17)**  & 0.60 (0.18)*** & 0.39 (0.18)*   & 0.52 (0.20)**  \\
		\textbf{Age (years)}                       & 0.01 (0.00)**  & 0.00 (0.00)**  & 0.01 (0.00)**  & 0.01 (0.00)*** \\
		\textbf{Gender (ref: Male)} &&&& \\
		\hspace{1em}Female                            & 0.00 (0.04)    & -0.03 (0.04)   & 0.00 (0.04)    & -0.02 (0.05)   \\
		\hspace{1em}Non-binary                        & -0.08 (0.12)   & 0.01 (0.13)    & -0.05 (0.13)   & -0.02 (0.14)   \\
		\textbf{Race (ref: White)} &&&& \\
		\hspace{1em}Asian                             & 0.12 (0.07)    & 0.03 (0.08)    & 0.06 (0.08)    & 0.02 (0.09)    \\
		\hspace{1em}Black or African American         & 0.21 (0.06)*** & 0.19 (0.06)**  & 0.22 (0.06)*** & 0.19 (0.07)**  \\
		\hspace{1em}Hispanic or Latino                & -0.08 (0.14)   & -0.03 (0.15)   & -0.15 (0.15)   & -0.07 (0.16)   \\
		\hspace{1em}Other race              & -0.11 (0.13)   & -0.23 (0.14)+  & -0.18 (0.13)   & -0.29 (0.15)*  \\
		\textbf{Occupation (ref: Employed) }&&&& \\
		\hspace{1em}Unemployed                        & 0.03 (0.07)    & -0.08 (0.07)   & 0.05 (0.07)    & -0.07 (0.08)   \\
		\hspace{1em}Homemaker                         & -0.08 (0.10)   & 0.01 (0.10)    & -0.05 (0.10)   & -0.05 (0.11)   \\
		\hspace{1em}Retired                           & -0.11 (0.07)   & -0.12 (0.08)   & -0.11 (0.08)   & -0.14 (0.09)+  \\
		\hspace{1em}Self-employed                     & 0.02 (0.06)    & 0.03 (0.06)    & 0.05 (0.06)    & 0.01 (0.07)    \\
		\hspace{1em}Student                           & 0.11 (0.08)    & 0.11 (0.09)    & 0.15 (0.09)+   & 0.16 (0.09)+   \\
		\hspace{1em}Other occupation                  & -0.07 (0.14)   & 0.04 (0.15)    & -0.04 (0.15)   & 0.01 (0.16)    \\
		\textbf{Education (ref: Less than high school)} &&&& \\
		\hspace{1em}High school diploma or equivalent & -0.04 (0.15)   & -0.08 (0.16)   & 0.02 (0.16)    & -0.07 (0.17)   \\
		\hspace{1em}College degree                    & 0.03 (0.15)    & -0.09 (0.16)   & 0.07 (0.16)    & -0.09 (0.18)   \\
		\hspace{1em}Graduate degree                   & 0.02 (0.16)    & -0.01 (0.17)   & 0.10 (0.17)    & -0.01 (0.18)   \\
		\hspace{1em}Other education                   & -0.03 (0.20)   & 0.04 (0.22)    & 0.00 (0.22)    & -0.03 (0.23)   \\ \cmidrule(){1-1}
		SD (instance intercepts)          & 0.48           & 0.43           & 0.43           & 0.41           \\
		SD (annotator intercepts)         & 0.22           & 0.26           & 0.25           & 0.29           \\
		Num.Obs.                          & 4500           & 4481           & 4500           & 4481           \\
		R2 Marg.                          & 0.012          & 0.014          & 0.015          & 0.018          \\
		R2 Cond.                          & 0.325          & 0.314          & 0.301          & 0.317          \\
		ICC                               & 0.3            & 0.3            & 0.3            & 0.3            \\
		\bottomrule
	\end{tabularx}
	\caption{Results of the analysis of individual prompts for the offensiveness rating task. This table displays coefficients for the two models (GPT-4o and Claude) and indicates the effects of specific socio-demographic attributes of human annotators on the discrepancies between human and LLM annotations. Positive coefficients signify that the LLM is less accurate in mimicking the responses of annotators from certain demographic categories.}
	\label{tab:varoff}
\end{table*}

\begin{table*}[t]
	\centering \small \setlength{\tabcolsep}{3pt}
	\begin{tabularx}{\linewidth}{X ll ll}
		\toprule
		& \multicolumn{2}{c}{Politeness Prompt 1} & \multicolumn{2}{c}{Politeness Prompt 2}  \\
		\cmidrule(r){2-3}\cmidrule(l){4-5}
		Socio-Demographic Attribute & GPT-4 & Claude & GPT-4  & Claude  \\
		\cmidrule(r){1-1} \cmidrule(lr){2-2} \cmidrule(lr){3-3} \cmidrule(lr){4-4} \cmidrule(l){5-5} 
		\textbf{Intercept}                         & 1.40 (0.19)*** & 1.43 (0.19)*** & 1.51 (0.19)*** & 1.65 (0.21)*** \\
		\textbf{Age (years)}                       & 0.00 (0.00)    & 0.00 (0.00)    & 0.00 (0.00)    & 0.00 (0.00)    \\
		\textbf{Gender (ref: Male)} &&&& \\
		\hspace{1em}Female                            & -0.05 (0.03)   & -0.05 (0.03)   & -0.05 (0.03)   & -0.05 (0.04)   \\
		\hspace{1em}Non-binary                        & -0.07 (0.10)   & -0.05 (0.10)   & -0.04 (0.10)   & -0.05 (0.11)   \\
		\textbf{Race (ref: White)} &&&& \\
		\hspace{1em}Asian                             & -0.08 (0.06)   & 0.00 (0.07)    & -0.08 (0.07)   & -0.02 (0.07)   \\
		\hspace{1em}Black or African American         & 0.13 (0.05)**  & 0.15 (0.05)**  & 0.15 (0.05)**  & 0.16 (0.05)**  \\
		\hspace{1em}Hispanic or Latino                & 0.07 (0.06)    & 0.14 (0.06)*   & 0.10 (0.06)+   & 0.11 (0.07)    \\
		\hspace{1em}Other race              & -0.23 (0.18)   & -0.15 (0.18)   & -0.11 (0.18)   & -0.15 (0.20)   \\
		\textbf{Occupation (ref: Employed)} &&&& \\
		\hspace{1em}Unemployed                        & -0.08 (0.05)   & -0.07 (0.05)   & -0.07 (0.05)   & -0.10 (0.06)+  \\
		\hspace{1em}Homemaker                         & -0.03 (0.07)   & -0.05 (0.07)   & -0.03 (0.07)   & -0.06 (0.07)   \\
		\hspace{1em}Retired                           & 0.04 (0.06)    & 0.04 (0.06)    & 0.02 (0.06)    & 0.03 (0.06)    \\
		\hspace{1em}Self-employed                     & -0.02 (0.05)   & -0.02 (0.05)   & -0.04 (0.05)   & -0.03 (0.06)   \\
		\hspace{1em}Student                           & -0.10 (0.08)   & -0.11 (0.08)   & -0.11 (0.08)   & -0.14 (0.08)+  \\
		\hspace{1em}Other occupation                  & 0.08 (0.11)    & 0.10 (0.11)    & 0.08 (0.11)    & 0.11 (0.12)    \\
		\textbf{Education (ref: Less than high school)} &&&& \\
		\hspace{1em}High school diploma or equivalent & -0.31 (0.18)+  & -0.35 (0.18)+  & -0.37 (0.18)*  & -0.40 (0.20)*  \\
		\hspace{1em}College degree                    & -0.41 (0.18)*  & -0.45 (0.18)*  & -0.46 (0.18)*  & -0.52 (0.20)** \\
		\hspace{1em}Graduate degree                   & -0.33 (0.18)+  & -0.41 (0.18)*  & -0.40 (0.19)*  & -0.49 (0.20)*  \\
		\hspace{1em}Other education                   & -0.46 (0.21)*  & -0.56 (0.21)** & -0.54 (0.21)*  & -0.60 (0.23)** \\
		\cmidrule(){1-1}
		SD (instance intercepts)          & 0.41           & 0.34           & 0.44           & 0.40           \\
		SD (annotator intercepts)         & 0.30           & 0.30           & 0.30           & 0.34           \\
		Num.Obs.                          & 11151          & 11151          & 11151          & 11151          \\
		R2 Marg.                          & 0.011          & 0.013          & 0.010          & 0.013          \\
		R2 Cond.                          & 0.322          & 0.292          & 0.340          & 0.340          \\
		BIC                               & 28301.7        & 27294.4        & 28833.6        & 28424.2        \\
		ICC                               & 0.3            & 0.3            & 0.3            & 0.3            \\
		RMSE                              & 0.67           & 0.66           & 0.67           & 0.67           \\
		\bottomrule
	\end{tabularx}
	\caption{Results of the analysis of individual prompts for the politeness rating task. This table presents the coefficients for the two models (GPT-4o and Claude), demonstrating how different socio-demographic characteristics of human annotators influence the distance between human and LLM annotations. Positive coefficients indicate a lower accuracy of the LLM in reflecting the views of annotators from particular demographic categories.}
	\label{tab:varpol}
\end{table*}

\subsection{Results for Placebo Prompting}
\label{sec:apppla}
\Cref{tab:mdistp_combined} presents the results for placebo prompting.
It includes sample sizes and mean distance scores ($\Delta_\mu$) for predictions generated using placebo prompts compared to those produced with no-info prompts for the models GPT-4o and Claude across two rating tasks.
The results show that placebo prompting (\Pprompts\ prompts) does not yield any notable differences relative to no-info prompting (\Nprompts\ prompts) for specific attribute values.
Overall, the scores remain stable across these comparisons.

\begin{table*}[t]
	\centering \small
	\begin{tabular}{ll rrr rrr}
		\toprule
		&&\multicolumn{3}{c}{Offensiveness}&\multicolumn{3}{c}{Politeness}\\
		\cmidrule(rl){3-5}\cmidrule(rl){6-8}
		\multicolumn{2}{l}{Placebo Attributes}& Count&$\Delta_\mu$ (GPT-4o)&$\Delta_\mu$ (Claude)& Count&$\Delta_\mu$ (GPT-4o)&$\Delta_\mu$ (Claude) \\
		\cmidrule(r){1-2}\cmidrule(rl){3-3}\cmidrule(lr){4-4}\cmidrule(lr){5-5}\cmidrule(rl){6-6}\cmidrule(rl){7-7}\cmidrule(r){8-8}
		\multicolumn{2}{l}{Height}&&&&&&\\
		&140 cm&601& 0.23& 0.19&1,380& 0.27& 0.19\\
		&150 cm&555& 0.27& 0.20&1,393& 0.27& 0.18\\
		&160 cm&579& 0.23& 0.19&1,417& 0.26& 0.20\\
		&170 cm&536& 0.20& 0.20&1,407& 0.27& 0.18\\
		&180 cm&574& 0.24& 0.20&1,354& 0.24& 0.18\\
		&190 cm&564& 0.28& 0.23&1,385& 0.25& 0.18\\
		&200 cm&572& 0.24& 0.24&1,446& 0.27& 0.18\\
		&210 cm&519& 0.25& 0.22&1,369& 0.25& 0.18\\
		\multicolumn{2}{l}{Zodiac sign}&&&&&&\\
		&Aries&395& 0.24& 0.22&889& 0.25& 0.17\\
		&Taurus&382& 0.26& 0.20&913& 0.27& 0.18\\
		&Gemini&355& 0.21& 0.23&895& 0.26& 0.18\\
		&Cancer&383& 0.29& 0.27&956& 0.27& 0.18\\
		&Leo&378& 0.24& 0.22&928& 0.27& 0.19\\
		&Virgo&360& 0.22& 0.23&915& 0.26& 0.19\\
		&Libra&398& 0.23& 0.19&974& 0.24& 0.19\\
		&Scorpio&390& 0.25& 0.18&975& 0.26& 0.19\\
		&Sagittarius&352& 0.23& 0.20&919& 0.27& 0.19\\
		&Capricorn&387& 0.25& 0.21&983& 0.27& 0.18\\
		&Aquarius&365& 0.28& 0.21&892& 0.25& 0.17\\
		&Pisces&355& 0.23& 0.17&912& 0.24& 0.20\\
		\multicolumn{2}{l}{House number}&&&&&&\\
		&6&460& 0.26& 0.21&1,130& 0.28& 0.19\\
		&12&446& 0.24& 0.22&1,090& 0.26& 0.18\\
		&13&447& 0.23& 0.21&1,107& 0.26& 0.19\\
		&24&424& 0.27& 0.22&1,103& 0.25& 0.18\\
		&45&455& 0.24& 0.18&1,123& 0.26& 0.19\\
		&68&438& 0.22& 0.19&1,098& 0.26& 0.18\\
		&98&456& 0.27& 0.23&1,190& 0.25& 0.20\\
		&122&465& 0.23& 0.21&1,116& 0.26& 0.19\\
		&234&466& 0.23& 0.21&1,118& 0.26& 0.19\\
		&1265&443& 0.26& 0.22&1,076& 0.26& 0.17\\
		\multicolumn{2}{l}{Enjoying}&&&&&&\\
		&food&1,468& 0.25& 0.22&3,793& 0.26& 0.19\\
		&sleep&1,486& 0.23& 0.21&3,662& 0.26& 0.18\\
		&friends&1,546& 0.25& 0.21&3,696& 0.26& 0.19\\
		\multicolumn{2}{l}{Favorite colour}&&&&&&\\
		&red&436& 0.25& 0.22&1,046& 0.28& 0.16\\
		&green&399& 0.27& 0.24&1,013& 0.26& 0.18\\
		&blue&400& 0.23& 0.21&997& 0.27& 0.19\\
		&yellow&444& 0.24& 0.22&984& 0.25& 0.19\\
		&purple&443& 0.27& 0.20&1,055& 0.25& 0.20\\
		&turquoise&367& 0.23& 0.20&1,056& 0.23& 0.21\\
		&orange&405& 0.23& 0.21&1,033& 0.26& 0.18\\
		&pink&371& 0.25& 0.21&976& 0.28& 0.16\\
		&black&441& 0.21& 0.20&993& 0.25& 0.17\\
		&white&381& 0.27& 0.21&991& 0.28& 0.20\\
		&brown&413& 0.21& 0.21&1,007& 0.26& 0.19\\
		\bottomrule
	\end{tabular}
	\caption{Sample sizes and mean distance scores of predictions for placebo prompting (\Pprompts\ prompts) predictions  in comparison to predictions with \Nprompts\ prompts for models GPT-4o and Claude at two rating tasks.}
	\label{tab:mdistp_combined}
\end{table*}

\end{document}